\documentclass[10pt,twocolumn,letterpaper]{article}

\usepackage{wacv}
\usepackage{times}
\usepackage{epsfig}
\usepackage{graphicx}
\usepackage{amsmath}
\usepackage{amssymb}

\usepackage{multirow}
\usepackage{subfig}
\usepackage{graphicx}
\usepackage{indentfirst}
\usepackage{arydshln}
\usepackage{booktabs}  
\usepackage{enumitem}
\setlist{nosep}

\wacvfinalcopy 

\usepackage[pagebackref=true,breaklinks=true,colorlinks,bookmarks=false]{hyperref}

\begin{document}

\title{Joint Pruning \& Quantization for Extremely Sparse Neural Networks}

\author{Po-Hsiang Yu\\
National Taiwan University\\
{\tt\small r08943024@ntu.edu.tw}
\and
Sih-Sian Wu\\
National Taiwan University\\
{\tt\small benwu@video.ee.ntu.edu.tw}
\and
Jan P. Klopp\\
National Taiwan University\\
{\tt\small kloppjp@gmail.com}
\and
Liang-Gee Chen\\
National Taiwan University\\
{\tt\small lgchen@ntu.edu.tw}
\and
Shao-Yi Chien\\
National Taiwan University\\
{\tt\small sychien@ntu.edu.tw}
}

\maketitle

\begin{abstract}
We investigate pruning and quantization for deep neural networks. Our goal is to achieve extremely high sparsity for quantized networks to enable implementation on low cost and low power accelerator hardware. In a practical scenario, there are particularly many applications for dense prediction tasks, hence we choose stereo depth estimation as target.

We propose a two stage pruning and quantization pipeline and introduce a Taylor Score alongside a new fine-tuning mode to achieve extreme sparsity without sacrificing performance. 

Our evaluation does not only show that pruning and quantization should be investigated jointly, but also shows that almost 99\% of memory demand can be cut while hardware costs can be reduced up to 99.9\%. In addition, to compare with other works, we demonstrate that our pruning stage alone beats the state-of-the-art when applied to ResNet on CIFAR10 and ImageNet.
	
\end{abstract}

\section{Introduction}
Deep learning based computer vision is becoming one of the foundations for autonomous agents and has a multitude of applications in augmented reality, wearable devices and other mobile platforms. All of these tasks require real-time capability of the underlying algorithms, which easily collides with the huge resource demands of deep neural networks. Integrated circuit designers have taken up this challenge and proposed ways to design efficient hardware implementations. The efficiency of those designs often hinges on characteristics of the neural network, for example sparsity. Sparsity is beneficial in two ways: it reduces the size of the network's parameters as well as the number of operations necessary. Previous research has proposed numerous methods to increase sparsity without sacrificing accuracy, but the symbiotic combination of pruning and quantization has only scarcely been discussed in the literature. 
\begin{figure}[t!]
	\centering
	\scriptsize
	\begin{tabular}{cc}
		Original PSM-Net~\cite{PSMNet} & Prune + Quantize (Weight Bit = 5) \\
		Model Memory Size: 18.51MB  & Model Memory Size: 0.39 MB \\
		Parameter : 5.2 M & Parameter : 0.1096 M \\
		\includegraphics[scale=0.0860]{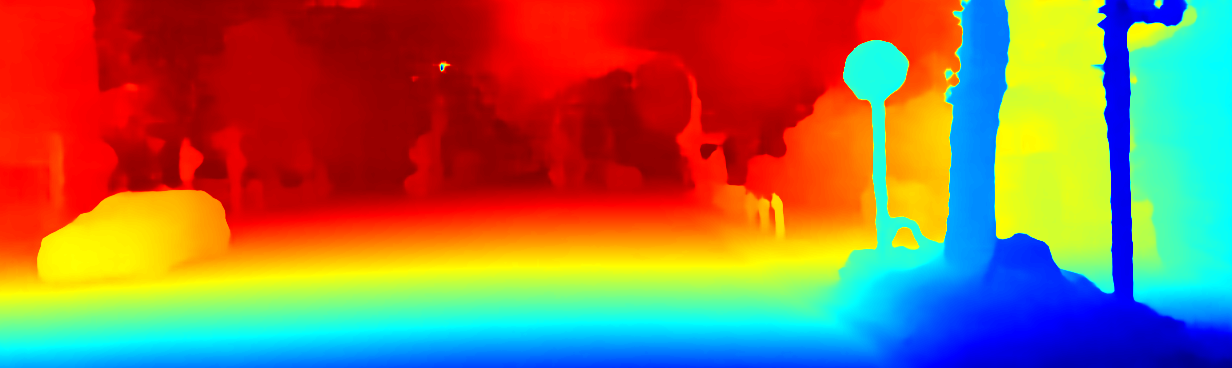} & \includegraphics[scale=0.0860]{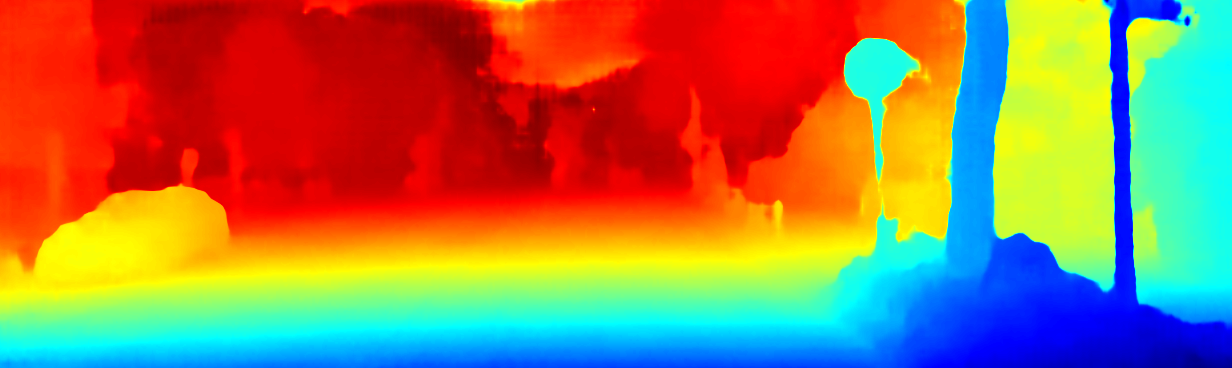}\\
		\includegraphics[scale=0.0860]{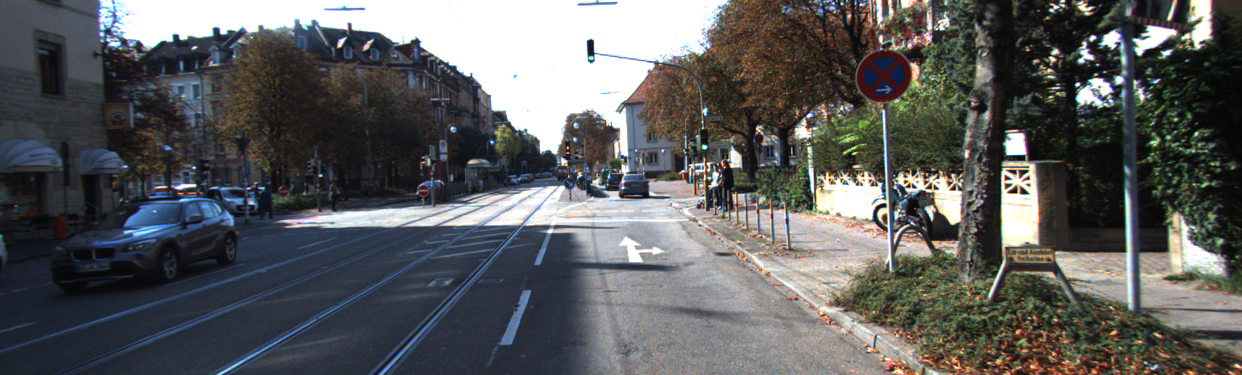} & \includegraphics[scale=0.0862]{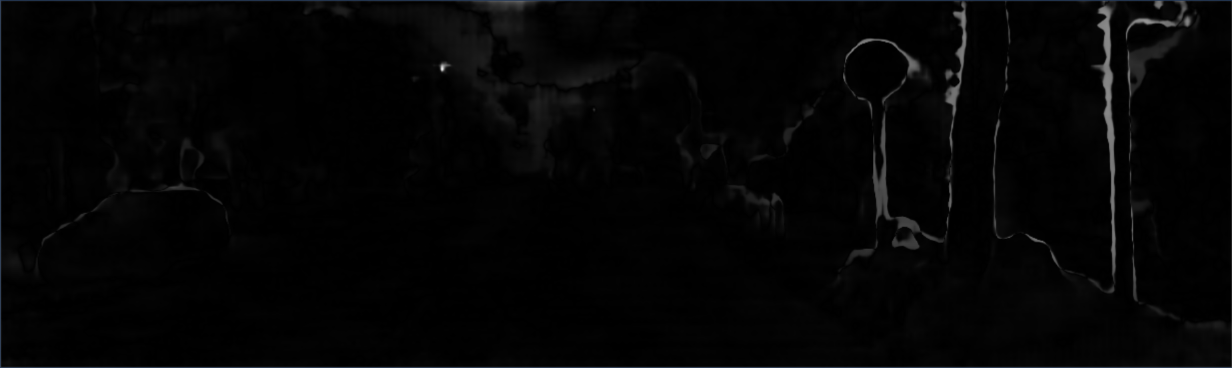}\\
	\end{tabular}
	\caption{Result of our pipeline. Upper row: depth predicted by the original and by the sparse \& quantized network. Bottom row: input image and difference between both depth predictions. (brightness indicates difference)}
	\label{fig:diff_WB2}
\end{figure}

Our work attempts to fill this gap. Using a two-stage pipeline with an adapted pruning criterion and a new fine-tuning mode, we extensively evaluate the influences of the pruning and the quantization stage. Different from previous works, we choose stereo depth matching as computer vision task to be optimised. The reason is that this dense regression problem is closer to real-world problems that require real-time computation on mobile devices. This real-time requirement also means that batch-processing is not possible. Hence, if the weights of the network cannot be stored on chip (see Table~\ref{table:stereoNN} for memory requirements and on-chip availability), they have to be loaded into the chip for every single processed sample, dramatically increasing bandwidth and power consumption. Therefore, decreasing the network's size can have an immediate impact on demand-sensitive mobile platforms. 

Overall, our contributions are as follows:
\begin{itemize}[leftmargin=*]
	\item A two-stage pruning and quantization pipeline with an adapted criterion and a novel fine-tuning strategy is proposed to minimize memory size and hardware cost.
	\item Besides demonstrating that the number of weights can be significantly reduced on stereo depth estimation networks, our pruning alone sets a new state-of-the-art for ResNet on both CIFAR10 and ImageNet.
	\item We show that pruning before quantization not only can increase sparsity, but also accuracy, because pruned weights cannot induce quantization noise at a later stage. We also show that interweaving pruning and quantisation can increase both performance measures even further. 
\end{itemize}

\begin{table}[t]
\centering
\footnotesize
\caption{(a) Computational complexity and parameters of existing stereo neural networks and (b) specifications of state-of-the-art neural network accelerators. The image resolution is full-HD at 60 fps and the disparity range is 512. }
\label{table:stereoNN}
\subfloat[State-of-the-art Stereo Neural Networks]{
\begin{tabular}{cccc}
\toprule
 & TOPs & Parameters (M) & Memory Size (MB) \\ \midrule
PSMNet~\cite{PSMNet} & 243.04 & 5.2 &   18.5  \\ \midrule
CSPN~\cite{CSPN} & 527.95  & 256  &   1086.0  \\ \bottomrule
\end{tabular}}

\subfloat[State-of-the-art Neural Network Accelerators]{
\begin{tabular}{ccc}
	\toprule             
AI Engine & TFLOPs & On-chip memory (MB)  \\ \midrule
GANPU~\cite{GANPUISSCC2020} & 24.13 & 0.66      \\ \midrule
MediaTekDLA~\cite{MediaTekDLA} & 3.52 & 2.125    \\ \bottomrule
\end{tabular}}
\end{table}

\section{Related Works}  
\subsection{Pruning Techniques}
The goal of pruning is to increase sparsity while maintaining performance as good possible. It can target single weights (unstructured) or entire filters, residual connections, or other building blocks of networks (structured).

Early works explored simple strategies to estimate importance. Han et al.~\cite{han2015learn,han2015deep} removes connects where weights are below a certain threshold, Li et al.~\cite{li2016pruningfilter} did this for entire filters based on their $L1$-norm. 
Luo et al.~\cite{luo2017thinet} improved this approach by pruning filters based on statistics from the succeeding layer. 
He et al.~\cite{he2018SoftFP,he2019GM} proposed soft-pruning where a pruned filter can recover as well as a new criterion for filter selection based on the geometric median.
Molchanov et al.~\cite{molchanov2016pruning,Importance} incorporated the model's loss function into the pruning decision via Taylor expansion both for channels (\cite{molchanov2016pruning}) as well as for individual weights (\cite{Importance}).
He et al.~\cite{he2020LFPC} developed a differentiable pruning criteria sampler to adaptively select different pruning criteria (including $L1$-norm, $L2$-norm and geometric median) for different layers. Guo et al.~\cite{guo2020dmcp} modelled channel pruning as a Markov process in which each state represented for retaining the channel and transition between states denoted the pruning process. Luo et al.~\cite{luo2020residual_limited} focused on pruning residual connections via a KL-divergence based criterion and refined labels to prune with limited-data. Chin et al.~\cite{chin2020towards} proposed to learn a global ranking of the filter across different layers of the CNN, which alters the goal of model compression to producing a set of CNNs with different accuracy and latency trade-offs to speed up the pruning process.

\subsection{Quantization Techniques}
Similar to pruning, the motivation behind quantization is twofold: lower precision weights do obviously save space but their computation become easier as well, especially multiplication (which is roughly quadratic in the number of bit).

Han et.al.~\cite{han2015deep} presented weight quantization based on code-books after pruning in a three-stage pipeline with the sole goal of model compression by exploiting dense, fully-connected layers. 
To reduce computational complexity, some approaches focus on low precision weights: expectation back propagation (EBP)~\cite{ebp} shows how to backpropagate through discrete weights to ensure high accuracy after quantization. Courbariaux et.al.~\cite{binaryconnect} expanded on EBP by using full-precision weights as reference when performing weight binarization. Several works added low precision quantization for activations, with different precisions: the quantized neural network~\cite{hubara2016quantized}, the binary net~\cite{courbariaux2016binarized}, the ternary net ~\cite{li2016ternary}, as well as the XNOR-Net~\cite{rastegari2016xnor}. 
Zhou et.al.~\cite{INQ} introduced incremental network quantization (INQ) to convert full-precision weights into powers of two so that only shifters but no multipliers are required.
Beyond simply quantizing the network, some works aim at accelerating model inference:
Jacob et.al.~\cite{TFLite} quantized weights such that inference can be carried out by integer-only arithmetic while a training procedure was co-designed to preserve accuracy after quantization. 
Zhuang et.al.~\cite{zhung2018} proposed a two-stage optimization strategy to quantize weights and activations. 
\subsection{Discussion}
Quantization and pruning techniques are numerous, yet they are almost exclusively evaluated on sparse prediction tasks and their associated network architectures such as VGG~\cite{vgg} and ResNet~\cite{resnet, resnetv2} for CIFAR-10~\cite{cifar10} or ImageNet~\cite{imagenet}. Quantization and pruning for dense prediction networks is still quite unexplored.
Most of the prior works considered pruning from an algorithmic perspective, not directly taking into account that hardware typically needs extreme degrees of sparsity to reap cost and power gains. In contrast, our approach removes more than 98\% of weights and quantizes the remaining weights into powers of two to enable the use of shifters instead of multipliers on hardware level, dramatically reducing hardware complexity.

Molchanov et al.'s method \cite{Importance} bears some resemblance to our pruning step. Hence, we would like to highlight some of the differences: (1) ~\cite{Importance} focuses on the importance of an entire convolutional filter, while our method proposes to create extremely high sparsity 
and focuses on removing individual weights. 
We simplified the summation of group contribution and gating layers by simply focusing on a single neuron's importance as described in Sec.~\ref{sec:pruning_criterion}. (2) ~\cite{Importance} needs averaging importance scores, selecting the number of mini-batches between pruning iterations and choosing number of neurons to be pruned, which leads to a complex hyper-parameter setting and increases the difficulty to find the best configuration. 
In contrast, our method only requires a single threshold to determine the unimportant weights and the training will automatically converge to the highest sparsity possible under that threshold. 

Han et al.~\cite{han2015deep} also explore network compression via pruning and quantization. However, they use a different pruning technique ($L1$ norm), which has been outperformed by other methods since. Furthermore, they quantize via weight sharing. Also, their target is different from ours: we focus on dense prediction applications that use only convolutional layers while their compression gains are mainly contributed by fully-connected layers. 

\begin{figure*}[ht!]
\centering
\includegraphics[scale=0.29]{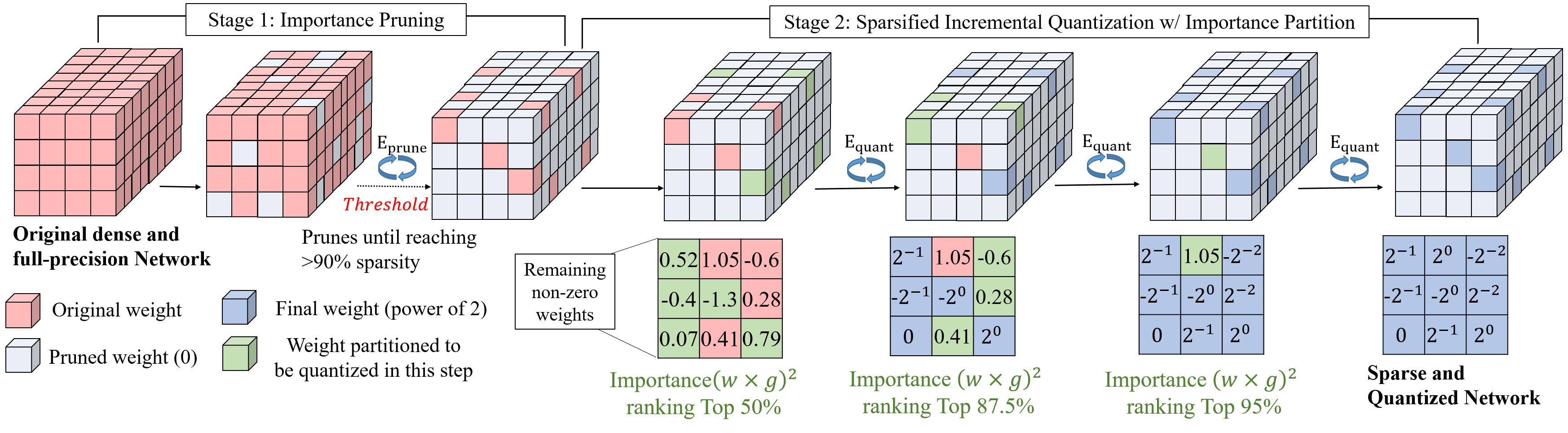}
\caption{Our two stage prune-then-quantize pipeline. Stage one provides sparsity above $90\%$ controlled by an importance threshold. Stage two iteratively quantizes the remaining non-zero weights into powers of two. It uses the same importance metric as stage one. In the figure, the quantization step (i.e. the relative amount of weights that are quantized) during stage two is set to $[0.5, 0.875, 0.95, 1]$ for simplicity, resulting in three iterations (no fine-tuning after the last quantization). 
}
\label{fig:SystemDiagram}
\end{figure*}

\begin{figure}[h!]
\centering
\includegraphics[scale=0.4]{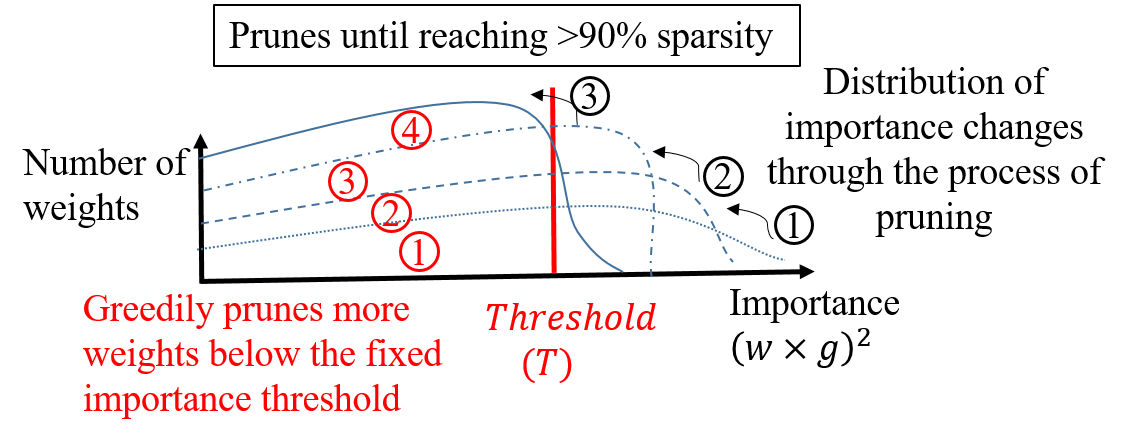}
\caption{Illustration of the importance threshold. At the beginning, the importance distribution of the network is as \textcircled{1}. During the pruning process, the fixed threshold gradually regularize the network importance distribution to shift to left \textcircled{2}, \textcircled{3}. By iteratively removing the weight below the threshold, a highly sparse network as \textcircled{4} is obtained.
}
\label{fig:threshold}
\end{figure}
\section{Proposed Method}

Our technique consists of two steps and aims at reducing memory and computation demands. Step 1 recursively prunes a pre-trained model to achieve up to 90\% weight sparsity and step 2 quantizes the remaining weights into powers of two. 
The system diagram is shown in Fig.~\ref{fig:SystemDiagram}. 

\subsection{Pruning Step}
Pruning aims at removing unimportant weights in each layer of the network. 
Stereo matching networks apply 3D convolution and pyramid pooling modules in their architectures (c.f. \cite{PSMNet, CSPN}), yielding large networks that are likely to contain many redundant parameters. 
The goal of the pruning step in our pipeline (left-hand side of Fig.~\ref{fig:SystemDiagram}) is to reduce this redundancy by setting unimportant weights to $0$.

\subsubsection{Pruning Criterion}
\label{sec:pruning_criterion}
If a parameter is important, the network's performance should drop significantly after removing it (setting it to 0). Considering a network with parameters $\mathbf{W}$ trained on a dataset $D$ and minimizing error $E$,
the squared difference of the errors $E$ with and without parameter $w_i$ is:
\begin{equation} \label{eq:3}
\Gamma = \left ( E ( D,\mathbf{W}) - E ( D ,\mathbf{W}|_{w_i}=0 ) \right )^2. 
\end{equation}
Computing Eq. ~\ref{eq:3} can be simplified by approximating it in the vicinity of the original parameters $\mathbf{W}$ using second-order Taylor expansion:
\begin{equation} \label{eq:4}
\Gamma^{(2)}(\mathbf{W})=(\mathbf{g} w_i - \frac{1}{2} w_i\mathbf{H_{m}W})^2. 
\end{equation}
where $\mathbf{H_{m}}$ is the Hessian of the network. $g_i$ are the elements of the weight gradient ($\frac{\partial E}{\partial w_i} $) and are readily available from back-propagation. We use an even simpler version of this approximation, retaining only the first term of the Taylor expansion:
\begin{equation} \label{eq:5}
\Gamma^{(1)}(\mathbf{W})=(g_i w_i)^2. 
\end{equation}
We define the \textbf{Taylor Score} $S_\text{Taylor}$ of a single weight as its gradient times it weight value according to Eq. ~\ref{eq:5}. This importance score is used to decide which weight should be removed during fine-tuning. This was inspired by Molchanov et al.'s work ~\cite{Importance}, but we simplified the complex summation of the group contribution and the gating layers by simply focusing on a single neuron's importance.  
If the importance score $S_\text{Taylor}(w_i)$ of a weight $w_i$ is smaller than a pre-defined threshold $T$, the weight would be set to $0$, thereby removing it from the computational graph. Fig.~\ref{fig:threshold} illustrates how the importance threshold reshapes the weight importance distribution during the pruning and fine-tuning cycles.

\subsubsection{Fine-tuning Strategy}\label{sec:finetuning_strategy}
During the course of fine-tuning, we greedily remove the weights whose $S_\text{Taylor}$ are smaller than the pre-defined threshold.
Since the distribution of weight importance scores varies during fine-tuning, the portion of parameters being pruned is different in each iteration. 
Thus, adopting the commonly used "soft" pruning where weights removed have a chance to recover in later iterations is not reasonable here: once pruned parameters recover, they may cross the importance threshold again and reduce sparsity. 
Instead, we adopt two fine-tuning strategies to retain accuracy while boosting sparsity.

\begin{figure}[h!]
	\centering
	\includegraphics[scale=0.22]{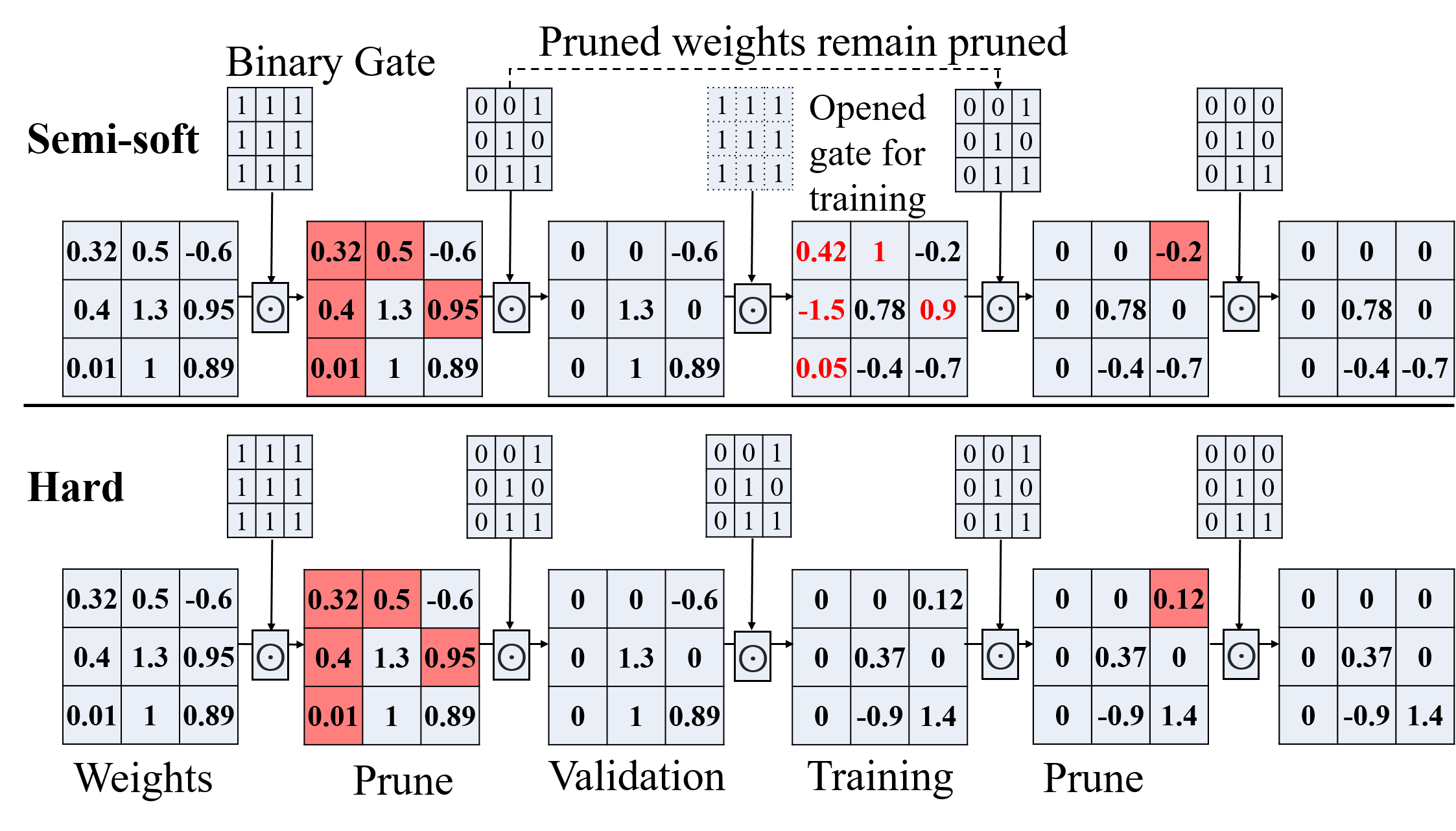}
	\caption{
		One iteration of two different fine-tuning strategies. 
		The upper one is semi-soft and the lower one is hard. 
		Red grids are the weights to be pruned. Pruned weights being updated during training are colored in red. 
		The $\odot$ operator represents Hadamard product. 
	}
	\label{fig:semisoft_hard}
\end{figure}

\textbf{Semi-soft Fine-tuning.}
We introduce binary gates to the network to control the presence of a weight.
Binary gates would be placed in front of each weight, where $1$ indicates the weight is used and $0$ means it is zero. 
If the weight is pruned away, its gate is set to $0$, otherwise, it remains at $1$.
The gate acts as a switch and plays different roles in training and testing. 
During \textbf{testing}, gates remain in their state so that pruned weights are zero and the sparse network is used. 
During \textbf{training}, no matter whether a weight has been pruned or not, the gate is in "$1$" state. 
This allows the original network architecture to be preserved which helps fine-tune the network as a whole.
However, once a gate has been set to "$0$", it will remain "$0$" testing. 
In other words, weights that have already been pruned can change but cannot be added back to the network. This is contrary to the common "soft" pruning in which the pruned-away weight may have a chance to recover in later fine-tuning stages. Hence, we denote it \textbf{semi-soft fine-tuning}. One iteration of fine-tuning is depicted in the upper row of Fig~\ref{fig:semisoft_hard}.

\textbf{Hard Fine-tuning.}
Another way to fine-tune after pruning is to only update the remaining parameters.
The gradients of the pruned parameters remain zero once they have been pruned.
As a result, the pruned parameters have no effect on the loss and back-propagation is carried out only through the remaining parameters, which are also the only ones being updated. 
An iteration of this variant is depicted in the bottom row of Fig~\ref{fig:semisoft_hard}. 

\subsubsection{Complete Pruning Algorithm}
A complete run of pruning consists of the following operations:
(1) obtain the gradient of each trainable parameter, 
(2) set the gate value of those parameters whose importance is below a pre-defined threshold to $0$, and 
(3) update either the whole network (semi-soft) or only the remaining parameters (hard), depending on the fine-tuning strategy chosen.
The whole process proceeds until the desired sparsity is met or the weight sparsity converges.
By adjusting the threshold and the number of fine-tuning epochs, we can achieve arbitrary degrees of sparsity. 
\subsection{Quantization Step}
\label{sec:quantization_step}
A key insight of our work is that a highly sparse networks helps to retain the model's performance during the quantization stage. Section~\ref{sec:exp_prunequant_vs_quant} will show corresponding experimental results supporting this claim. The reason is that pruned weights don't cause quantization errors, they are 0 anyway. The much lower parameter count of the pruned model hence facilitates weight partitioning and quantization.
As a result, the performance impact of quantization is negligible. We quantize weights to powers of two instead of simply reducing the number of bits. This makes the resulting model even more hardware-friendly because it can use shifters instead of multipliers.

To achieve a fully-quantized model with weights being powers of two, our quantization stage is inspired by the three-step operation of incremental network quantization (INQ) ~\cite{INQ}. 
The three steps are: weight partitioning, group-wise quantization, and re-training. 
During weight partitioning, INQ divides the weights in each layer into two groups according to their absolute value. One group is quantized to a pre-defined power of two and the other is re-trained to retain the performance. 
These three steps proceed until all weights are quantized.
This quantization process maps each weight to a power of two. For more details, please refer to \cite{INQ}. 
The original INQ was designed for classification tasks and directly quantizes the network without pruning.
To quantize stereo matching networks efficiently and contain the performance drop, we make the following modifications: 

First, we use the Taylor score ($S_\text{Taylor}$) as partitioning criterion similar to the pruning stage. 
Taking a quantization step size of $0.5$ as an example, the neuron is allocated into the quantized group if its Taylor score is larger than the median of weights of the whole layer. 
The Taylor score turns out to be a better metric when quantizing a deep and over-parametrized network. The results in Section~\ref{sec:diff_imp_quant} support this observation.  
The original INQ adopt random and absolute value partitioning, which fails to match our performance on larger stereo network.

Second, we add our pruning technique to every fine-tuning while performing incremental quantization. We incorporate pruning into the re-training step of quantization by removing remaining non-power-of-two weights whose $S_\text{Taylor}$ is below threshold $T$. This leaves less weights to be partitioned in the next quantization step. 
The sparsity can be further increased while sacrificing only little performance. Detailed results are presented in Section~\ref{sec:more_exploration}.
This highly enhances the flexibility of stereo matching network quantization. Performing quantization only, one cannot directly control the resulting sparsity. With our proposed method, the sparsity after quantization can be easily adjusted by simply incorporating pruning and changing the pruning threshold during quantization. 

\section{Experiments}
\subsection{Experimental Setting}
Our experimental evaluation is carried out on two end-to-end state-of-the-art stereo matching networks, PSM-Net~\cite{PSMNet} (KITTI 2015 ~\cite{KITTI2015C, KITTI2015J} dataset) and CSPN ~\cite{CSPN} (NYU V2~\cite{nyuv2} dataset). We use the implementations and pretrained models provided by the respective authors.
PSM-Net contains about $5$ million parameters while CSPN contains more than $200$ million.
Since referenced datasets are different, we use different performance metrics for KITTI2015 and NYU V2, respectively. 
For PSM-Net on KITTI2015, this is the 3-px error, which defines error pixels as those having end-point error greater than $3$. 
The 3-px accuracy is simply the 3px error subtracted from $100\%$.
The performance metric for CSPN on NYU V2 is $\delta_{1.02}$, which is calculated as the percentage of pixels whose predicted disparity does not deviate more than $2\%$ from the ground truth disparity. 
Thus, the 3-px error is the lower the better and $\delta_{1.02}$ is the higher the better.
During the pruning phase, we prune the network for $50$ epochs ($E_{\text{prune}}$), while at the quantization phase, we fine-tune the network for $3$ epochs ($E_{\text{quant}}$) during each quantization step. 
If not indicated otherwise, the Taylor Score threshold $T$ is $10^{-11}$ and the fine-tuning strategy is "Hard".
\subsection{Experimental Results of Prune-then-Quantize}
\begin{figure}[h!]
\centering
\includegraphics[scale=0.4]{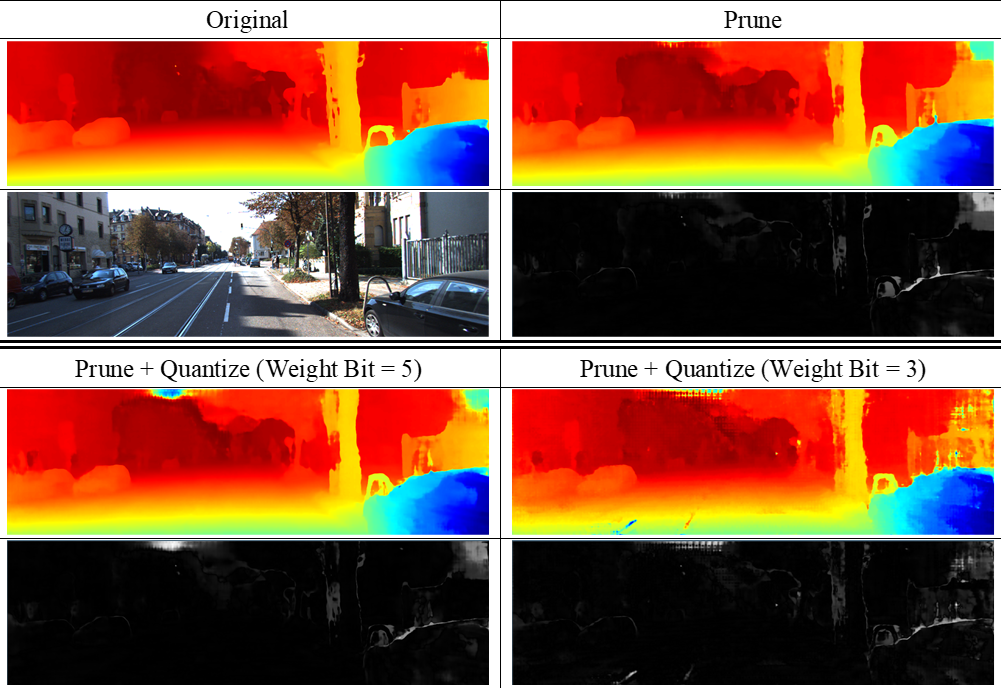}
\caption{Qualitative Results of PSM-Net}
\label{fig:qualitative}
\end{figure}

Qualitative experimental results for PSM-Net on a real scene extracted from the KITTI2015 dataset are shown in Fig. ~\ref{fig:qualitative}. 
The upper image of each of the pairs shows the disparity map obtained in different settings (original model, pruned model, pruned and quantized to 5 weigh bits, pruned and quantized to 3 weight bits). 
The bottom image shows the disparity difference between the original model and the predicted results. 
Brighter regions indicate larger differences.
We can observe that our pipeline hardly causes any errors to the disparity prediction. Most regions, including large, flat surfaces and complicated, detailed parts (eg. cars and trees), remain intact.  

Table~\ref{table:systemresult} lists accuracy, weight sparsity and memory reduction of original (O) and processed stereo neural networks in two configurations (A: high sparsity with quantized power of two weights, B: medium sparsity).
The two networks achieve weight sparsities of above 98\% and almost 94\% while sacrificing 2\% and 3\% in accuracy, respectively. In the medium sparsity configurations, accuracy is almost maintained (less then 1\% degradation), while sparsity is still high with 75\% and 82\%, respectively. 

With the remaining non-zero weights all being powers of two, not only are far less operations necessary but all multiply-accumulate operations can be replaced by arithmetic shifts. From a hardware cost perspective, a 16bit MAC operation can be realized by a 16bit hardware multiplier and a final accumulation, which can be further decomposed into 17 adders and 16 shifting operation. Thus, we approximate the cost of a shifting operation as $2/33$ of the cost of a normal MAC operation. The last column of Table~\ref{table:systemresult} shows a significant reduction of hardware cost achieved by our pruned and quantized network.
To verify potential memory reduction, we compressed the network parameters using Zip (Memory column). Above 98\% of memory reduction can be achieved with our prune-then-quantize method for both networks.

\begin{table}[h!]
\centering
\scriptsize
\caption{Sparsity, accuracy, model memory size and hardware cost of the proposed system. Configuration "A" has been processed by pruning and quantization (weight bit = 3). Configuration "B" has been pruned to show a different sparsity-accuracy trade-off, hence hardware costs equal TOPs. 
The TOPs are calculated under the assumption that the network runs at 60fps and Full HD resolution input. An operation is defined as a 16bit MAC and shift. O is the original network.}
\label{table:systemresult}
\begin{tabular}{lrrrrrrr}
	\toprule
	\multicolumn{7}{c}{PSM-Net~\cite{PSMNet}} \\ \midrule
        & 3-px Acc. & Spar.  & Para.(M) & Memory(MB) & TOPs & *Cost  \\ \midrule
O   & 99.10\% & 0.00\%  & 5.22 & 18.51 (100.0\%) & \multicolumn{2}{c}{243.04}\\ \midrule
A  & 97.14\% & 98.18\%   & 0.095 & 0.20 (1.1\%)& 1.01 &0.061\\ \midrule
B  & 98.54\% & 75.24\%   & 1.29 & 5.77 (31.2\%)& \multicolumn{2}{c}{54.82} \\ \midrule
\multicolumn{7}{c}{}  \\
\multicolumn{7}{c}{CSPN~\cite{CSPN}} \\ \midrule
        & $\delta_{1.02}$ & Spar.  & Para. (M) & Memory(MB) & TOPs & *Cost \\ \midrule
O   & 83.44\% & 0.01\%  & 256 & 1086 (100.0\%)& \multicolumn{2}{c}{527.85}\\ \midrule
A  & 80.69\% & 93.73\%  & 16 & 18.5 (1.7\%)& 38.23  & 2.32\\ \midrule
B  & 82.57\% & 81.95\%  & 46.2 & 176.2 (16.2\%)&  \multicolumn{2}{c}{88.08}\\ \bottomrule
\end{tabular}
\\ *The MAC harware cost is the same as operation, while the harware cost of shifting operation is calculated as $2/33$ of the cost of a normal MAC operation.
\end{table}

\subsection{Pruning Criterion and Fine-tuning Strategy} 
\begin{figure}[h!]
	\centering
	\includegraphics[scale=0.25]{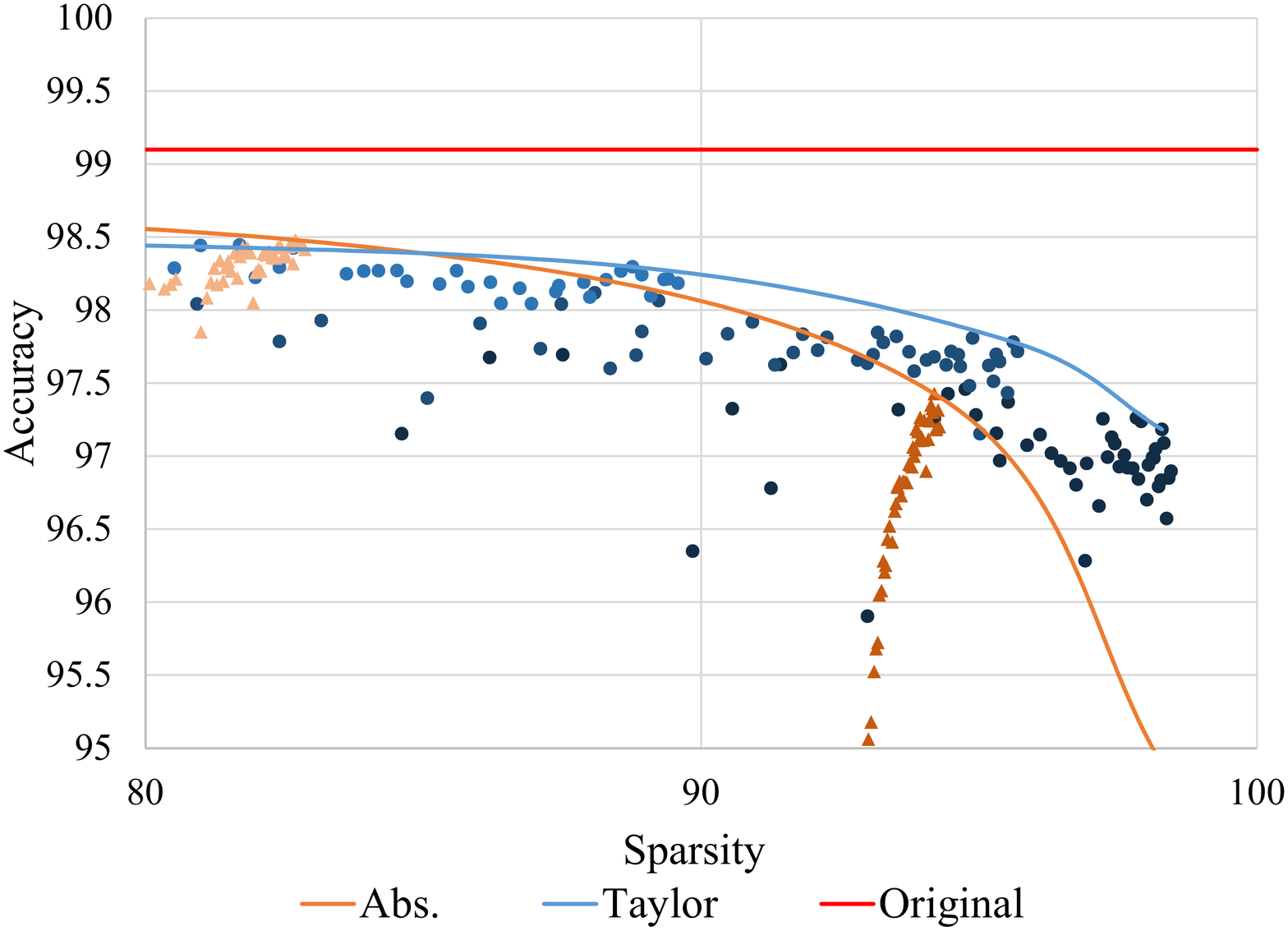}
	\caption{Dots in blue shades are $S_\text{Taylor}$ results after each epoch, while triangles in orange shades are results for the absolute value criterion. Darker dots mean higher threshold. $S_\text{Taylor}$ can achieve higher accuracy at sparsities above 85\%, which is the main focus of our work. Especially for a sparsity above 97\%, our pruning method is performing favourably. The graph covering all sparsity range can be refer to Fig~\ref{fig:criteria} in appendix.}
	
	\label{fig:criteria_2}
\end{figure}

The experiments in this section are conducted using PSM-Net trained on KITTI2015 with our proposed pruning technique. We compare a weight's absolute value and its Taylor Score at different importance thresholds.
As our goal is high sparsity, a suitable criterion should help retain accuracy even at sparsity levels above $90\%$. 
Fig.~\ref{fig:criteria_2} shows the results.  
In terms of accuracy, when the sparsity is above $85\%$, our Taylor Score outperforms the absolute value. It achieves a test accuracy of $97.7\%$ at $94\%$ sparsity, compared to $97.2\%$ at the same sparsity for the absolute value. 

 Fig.~\ref{fig:finetune_str} compares the two fine-tuning strategies mentioned in Section~\ref{sec:finetuning_strategy}. 
With the exception of the first few epochs, the "Hard" pruning scheme performs better than the "Semi-soft" one. This remains the case even for a lower threshold (i.e. less pruning) where accuracy is almost preserved.

When the sparsity gradually increasing, updating an already pruned weight may cause unstable dynamics, which decreases the accuracy. To achieve our target of removing $95\%$ of all parameters, \textbf{Hard} fine-tuning with high sparsity is a better strategy, while \textbf{Semi Soft} fine-tuning can be applied to medium sparsity cases. 
\begin{figure}[h!]
\centering
\includegraphics[scale=0.45]{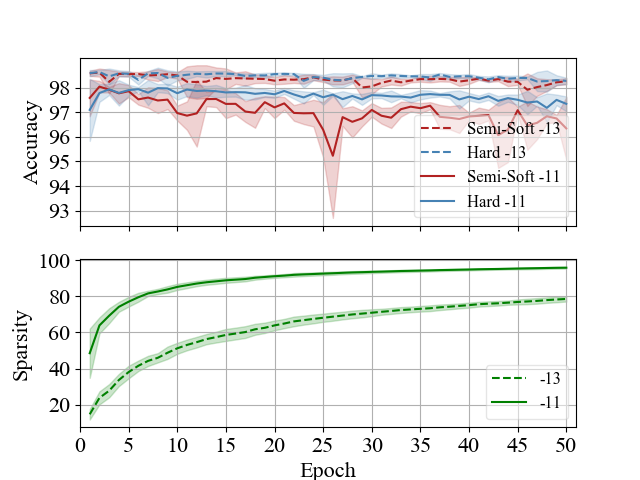}
\caption{Comparison of different fine-tune strategies. Lines represent the average over five experiments and the shaded area are the upper and lower bound of the standard deviation. "-11" and "-13" are two different pruning thresholds, 1E-11 and 1E-13, respectively. Only in the very beginning of the pruning process is "Semi-soft" pruning advantageous, afterwards "Hard" does not only dominate but also converges more stably. 
}
\label{fig:finetune_str}
\end{figure}

\subsection{Prune-then-Quantize vs. Quantization Only}
\label{sec:exp_prunequant_vs_quant}
\begin{figure}[h!]
	\centering
	\includegraphics[scale=0.385]{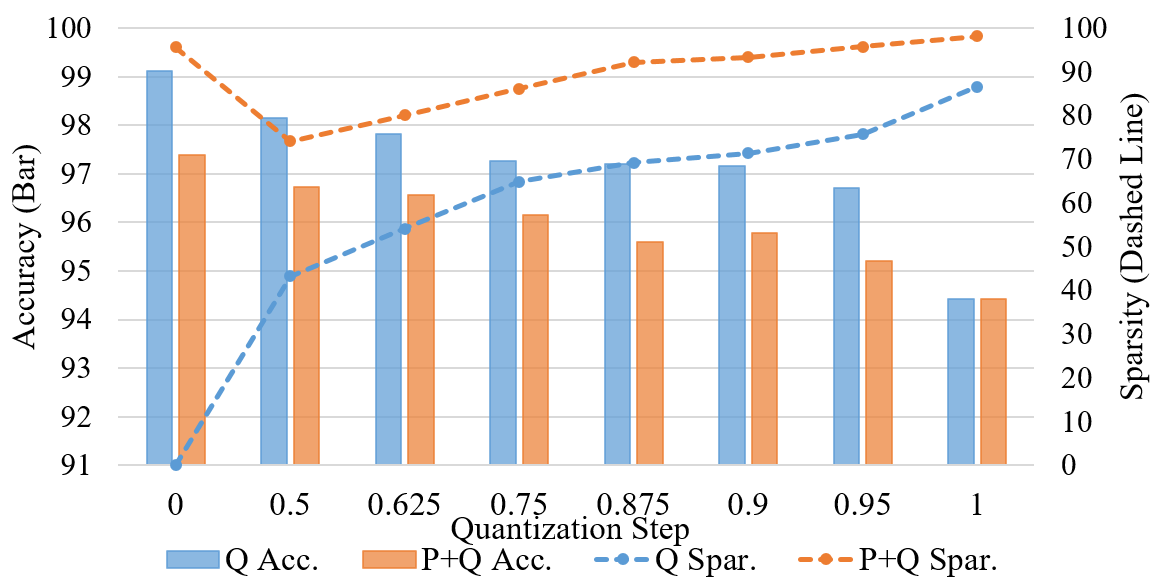}
	\caption{Effect of Sparse Pruned Model. "Q" stands for quantization only and "P+Q" for pruning and then quantizing. The bars represent accuracy and the lines represent sparsity after each quantization step. Intermediate models are only partially quantized, only the model at quantization step 1 is fully quantized. One can easily observe that our "prune the quantize" strategy achieves an accuracy similar to "quantization only" but retains only 1.8\% of the weights, compared to 13.4\%.}
	\label{fig:P_P+Q}
\end{figure}

We compare our prune-then-quantize pipeline to a quantization-only approach in terms of accuracy and sparsity.
The results in Fig.~\ref{fig:P_P+Q} show that by using our two-stage pipeline (using a pruned, high sparsity model for quantization), we can achieve much higher sparsity at less performance degradation.
By pruning first, neurons that contribute less to the output are set to zero, leaving only less than $5\%$ of the "important" neurons in the network. 
This eliminates the interference of those unimportant neurons when floating point numbers are mapped to quantized values as any weight already being zero cannot introduce a quantization error. As a result, for the same accuracy, our "prune then quantize" scheme retains only 1.8\% of the weights, while quantization only ends up requiring 13.4\%,  more than $7\times$ the amount.

\subsection{Different Importance Metrics for Quantization}
\label{sec:diff_imp_quant} 
We tested different importance metrics for weight partitioning in the quantization phase. 
 Table ~\ref{table:diff_metric} shows the achieved weight sparsity for different input neural networks.
There are fewer parameters left for $S_\text{Taylor}$ partitioning in each layer of PSM-Net, so the Abs. partition performs slightly better. 
Otherwise, for the much larger CSPN network, despite already highly sparsified layers, the number of parameters is still more than $40$ times larger than PSM-Net. 
The much higher parameter count provides sufficient weights to make $S_\text{Taylor}$ more informative. The weight with significant importance can be turned into a power of two first, which helps to build a more solid quantization process and obtain a fully quantized model with less performance loss.
\begin{table}[h!]
	\centering
	\footnotesize
	\caption{Sparsity and accuracy of different partition metrics on different input neural network.}
	\label{table:diff_metric}
	\centering
	\begin{tabular}{cccc}
		\hline
		&      &  Abs.                       & Taylor                     \\ \hline
		\multirow{2}{*}{PSMNet} & Acc.  &  \multicolumn{1}{c}{97.13} & \multicolumn{1}{c}{97.08} \\ \cline{2-4} 
		& Spar. &  \multicolumn{1}{c}{95.55} & \multicolumn{1}{c}{95.54} \\ \hline
		\multirow{2}{*}{CSPN}   & $\delta_{1.02}$                      & 0.8017                     & \textbf{0.8069}                     \\ \cline{2-4} 
		& Spar.                      & 93.79                      & 93.73                      \\ \hline
	\end{tabular}
\end{table}

\subsection{Pruning during Quantization}
\label{sec:more_exploration} 
\begin{figure}[h!]
\centering
\includegraphics[scale=0.26]{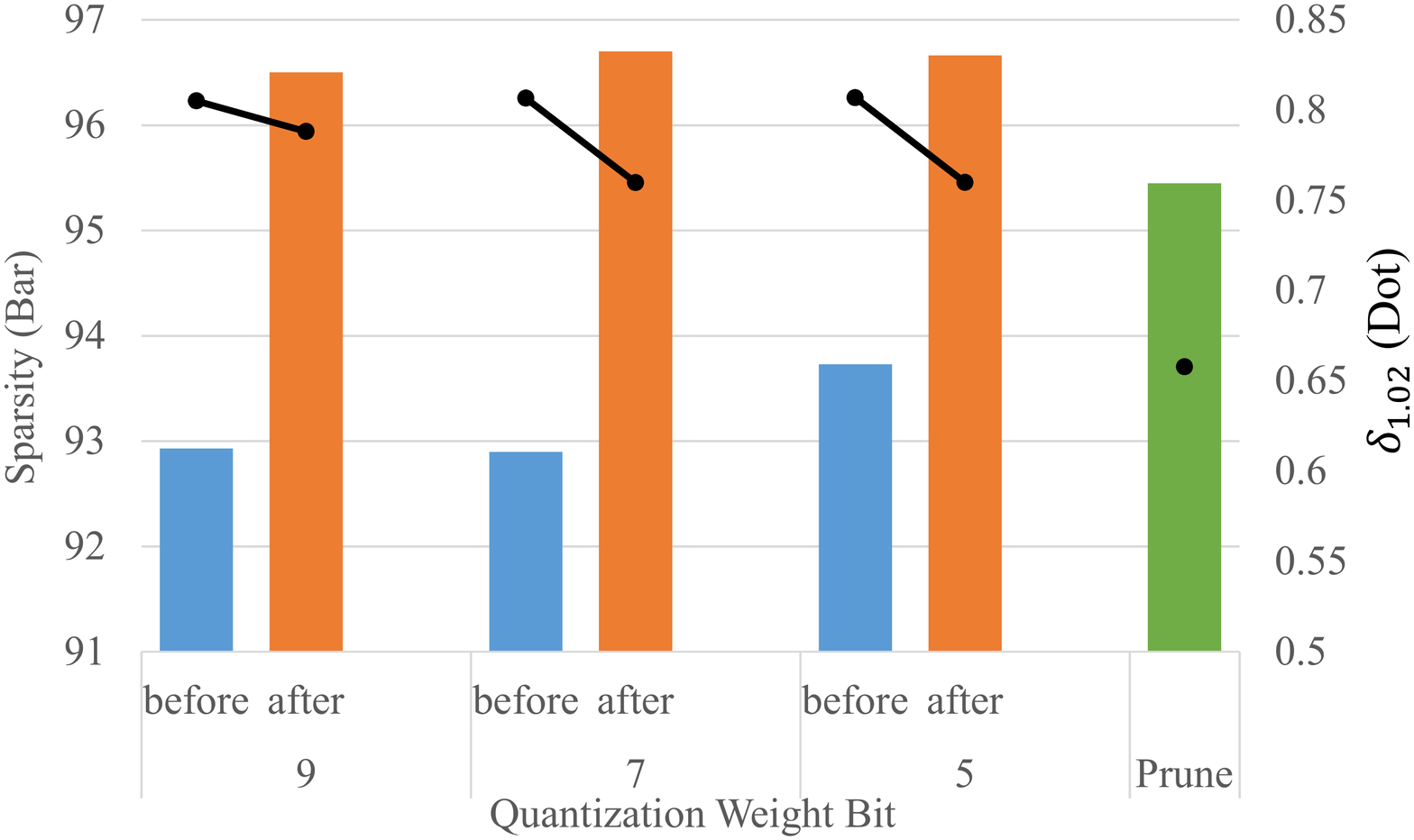}
\caption{Sparsity improvement results. Blue bars are w/o interleaved pruning. Orange bars prune during quantization. The green bar is the result of more aggressive pruning. Black dots show accuracy. By applying pruning during quantization, the sparsity can be further increased while performance drops moderately compared to only pruning more aggressively. This result underlines the insight that pruning and quantization should not be treated separately.}
\label{fig:spar_imp}
\end{figure}

As shown in previous experiments, the CSPN network's sparsity is still lacking behind at around 93\%(see Fig~\ref{fig:spar_imp} "before").
One could now move threshold $T$ to a higher value to capture more weights during the pruning stage. 
However, this would lead to severe accuracy deterioration (c.f. Fig~\ref{fig:spar_imp} right-most column) from about 0.81 down to 0.66.
To better maintain performance at high sparsity levels, we interweave pruning and quantization by continuing to prune during the fine-tuning phases of the quantization stage. 
As a result of this interleaved process, we can achieve a significantly higher sparsity than by simply increasing $T$ while the drop in accuracy is much smaller at the same time (see Fig~\ref{fig:spar_imp} "after"). 
The outcome of this interwoven strategy underlines once more that it is preferential to investigate pruning and quantization alongside each other.
\begin{table*}[h!]
	\centering
	\footnotesize
	\caption{Comparison of the pruned ResNet on ImageNet ILSVRC-2012. In "Method" column, Y and N indicate whether to use the pre-trained model to prune or not, respectively. In "Acc." column, the former is baseline accuracy and the latter is pruned accuracy. The "Acc.$\downarrow$" is the accuracy drop between pruned model and the baseline model, the smaller, the better.}
	\label{table:imagenet}
	\begin{tabular}{|c|c|c|c|c|c|c|c|}
		\hline
		Depth        & Method  & {Baseline Top1/ 5} & Top1(\%)   & Top1$\downarrow$(\%) & Top5(\%)   & Top5$\downarrow$(\%) & Sparsity \\ \hline \hline
		\multirow{4}{*}{18} & FPGM-mix~\cite{he2019GM}   & 70.28 / 89.63             & 68.41  & 1.87    & 88.48  & 1.15    & 28.10\%  \\
		& Ours\_1e-14\_39        & 69.76 / 89.08             & 69.59  & 0.17    & 89.1   & -0.02   & 30.60\%  \\
		& Ours\_1e-14\_100            & 69.76 / 89.08             & \textbf{70.15}  & \textbf{-0.39}   & \textbf{89.48}  & \textbf{-0.4}    & \textbf{47.58\%}  \\ 
		&Ours\_1e-13\_100            & 69.76 / 89.08             & 68.48  & 1.28   & 88.63  & 0.45& \textbf{69.95\%}  \\\hline \hline
		\multirow{7}{*}{34} & PFEC~\cite{li2016pruningfilter}                  & 73.23 / -                 & 72.17  & 1.06    & -      & -       & 10.80\%  \\
		& Taylor-FO~\cite{Importance}          & 73.31 / -                 & 72.83  & 0.48    & -      & -       & 18.00\%  \\
		& Ours\_1e-14\_14                & 73.31 / 91.42             & \textbf{73.04} & \textbf{0.27}   & \textbf{91.22} & \textbf{0.20}   & \textbf{21.27\%}  \\
		& FPGM-mix~\cite{he2019GM}          & 73.92 / 91.62             & 72.63  & 1.29    & 91.08  & 0.54    & 30.00\%  \\
		& Ours\_1e-14\_27       & 73.31 / 91.42             & \textbf{72.96} & \textbf{0.35}   & \textbf{91.23}  & \textbf{0.19}    & \textbf{31.08\%}  \\
		& Ours\_1e-14\_100              & 73.31 / 91.42             & \textbf{73.18} & \textbf{0.13}   & \textbf{91.26}  & \textbf{0.16}    & \textbf{55.65\%}  \\ 
		& Ours\_1e-13\_69              & 73.31 / 91.42             & 72.32 & 0.99   & 90.67  & 0.75    & \textbf{71.02\%}  \\ \hline \hline
		\multirow{9}{*}{50} & Taylor-FO~\cite{Importance}     & 76.18 / -                 & 74.50  & 1.68    & -      & -       & 28.00\%  \\
		& Taylor-FO~\cite{Importance}      & 76.18 / -                 & 71.69  & 4.49    & -      & -       & 44.00\%  \\
		& SFP~\cite{he2018SoftFP}           & 76.15 / 92.87             & 62.14 &  14.01  & 84.60 & 8.27 & 30.00\%  \\
		& FPGM-only~\cite{he2019GM}          & 76.15 / 92.87             & 75.59  & 0.56    & 92.23  & 0.24    & 30.00\%  \\
		& Ours\_1e-14\_22       & 76.13 / 92.87             & \textbf{75.71} & \textbf{0.42}   & \textbf{92.67}  & \textbf{0.20}    & \textbf{31.00\%}  \\
		& FPGM-only~\cite{he2019GM}          & 76.15 / 92.87             & 74.83  & 1.32    & 92.32  & 0.55    & 40.00\%  \\
		& Ours\_1e-14\_40               & 76.13 / 92.87             & 75.52 & 0.61   & 92.5  & 0.28    & \textbf{41.31\%}  \\
		& LFPC~\cite{he2020LFPC}              & 76.15 / 92.87             & 74.46 & 1.69   & 92.04  & 0.83    & *47.2\%  \\ 
		& Ours\_1e-14\_100               & 76.13 / 92.87             & 75.48 & 0.65   & 92.77  & 0.1    & \textbf{57.81\%}  \\\hline 
	\end{tabular}
	\\The "*" sign indicates that the sparsity is estimated.
\end{table*}
\subsection{Application to Classification}
To further validate the effect of our pruning method and compare it to other approaches, we applied our pruning technique to ResNet~\cite{resnet} on CIFAR-10~\cite{cifar10} and ILSVRC-2012~\cite{imagenet} datasets. We compare with the current state-of-the-art in network pruning, namely PFEC~\cite{li2016pruningfilter}, SFP~\cite{he2018SoftFP}, Taylor-FO~\cite{Importance}, and FPGM~\cite{he2019GM}. 

The training details are provided in the supplementary material. 
Experiments show that our pruning method achieves comparable performance. 
In all experiments we adopted \textbf{Hard Fine-tuning} and \textbf{Taylor Score}. By setting the importance threshold to be \emph{T} (in exponential notation) and the epoch after which the desired sparsity was obtained as \emph{E}, our results are indicated as Ours\emph{\_T\_E}.   

\textbf{CIFAR10.} For the CIFAR-10 dataset, we tested our pruning method on ResNet with depth $20$ and $56$. As shown in Table~\ref{table:cifar}, the results validates our method. For pruning both pretrained model and model trained from scratch, our method achieved superior accuracy under the same sparsity. As our method focuses on higher sparsity domain, we can even achieve above $60\%$ sparsity with the same or better accuracy compared with prior works.

\textbf{ILSVRC-2012.} Tabel~\ref{table:imagenet} shows that our pruning method outperforms other works on ILSVRC-2012. We can achieve higher sparsity with higher accuracy. The accuracy is retained even at above $55\%$ sparsity. For pruned pre-trained ResNet18, we can even achieve $70.15\%$ top-$1$ accuracy with $47.58\%$ sparsity. We believe that our pruning method may act as a regularizer when fine-tuning. Unimportant weights are removed, in this way, only the important weights are fine-tuned, which help regularize the training process.

\begin{table}[h!]
\centering
\scriptsize
\caption{Comparison of the pruned ResNet on CIFAR-10. The "$\downarrow$" has the same meaning as Table~\ref{table:imagenet}.}
\label{table:cifar}
\begin{tabular}{|c|c|c|c|c|c|c|}
\hline 
Depth                      & \multicolumn{2}{c|}{Method} & \multicolumn{2}{c|}{Acc. (\%)} & Acc.$\downarrow$(\%) & Sparsity \\ \hline \hline
\multirow{9}{*}{20} & Taylor-FO~\cite{Importance}   & Y          & 92.00                & 91.52     & 0.48      & 30.0\%  \\
                    & Ours\_1e-14\_61  & Y          & 92.45             & 92.12     & \textbf{0.33}      & 31.3\%  \\
                           & Taylor-FO~\cite{Importance}   & Y          & 92.00                & 89.78     & 2.22      & 65.0\%  \\
                           & Ours\_3e-13\_61  & Y          & 92.45             & 90.39     & \textbf{2.06}      & 65.7\%  \\ \cdashline{2-7}
                           & FPGM-only~\cite{he2019GM}   & N          & 92.20              & 91.09     & 1.11      & 29.2\%  \\
                           & Ours\_1e-15\_178 & N          & 92.45             & 92.18     & \textbf{0.27}      & 29.7\%  \\
                           & FPGM-mix~\cite{he2019GM}    & N          & 92.20              & 90.62     & 1.58      & 38.7\%  \\
                           & Ours\_5e-15\_161 & N          & 92.45             & 91.99     & \textbf{0.46}      & 41.0\%  \\
                           & Ours\_1e-12\_191 & N          & 92.45             & 90.88     & 1.57      & \textbf{63.2\%}  \\ \hline \hline
\multirow{9}{*}{56} & PFEC~\cite{li2016pruningfilter}           & Y          & 93.04             & 93.06     & -0.02     & 13.7\%  \\
                           & FPGM-only~\cite{he2019GM}   & Y          & 93.59             & 93.49     & 0.1       & 38.7\%  \\
                           & Ours\_1e-15\_36  & Y          & 94.09             & 93.66     & 0.43      & 40.9\%  \\
                           & Ours\_1e-14\_99  & Y          & 94.09             & 93.42     & 0.67      & \textbf{64.0\%}  \\ \cdashline{2-7}
                           & PFEC~\cite{li2016pruningfilter}           & N          & 93.04             & 91.31     & 1.73      & 13.7\%  \\
                           & SFP~\cite{he2018SoftFP}            & N          & 93.59             & 92.26     & 1.33      & 38.7\%  \\
                           & FPGM-only~\cite{he2019GM}   & N          & 93.59             & 92.93     & 0.66      & 38.7\%  \\
                           & Ours\_1e-16\_193 & N          & 94.09             & 93.72     & \textbf{0.37}      & 38.7\%  \\
                           & Ours\_7e-16\_198 & N          & 94.09             & 93.56     & 0.53      & \textbf{53.9\%}  \\ \hline
\end{tabular}
\end{table}

\section{Conclusion}
A prune-then-quantize technique is proposed to sparsify state-of-the-art stereo matching neural networks.
Our automatic pruning process is simple to use and requires only a single parameter to be set.
Following pruning, an incremental quantization method is adopted to convert the remaining weights into power of two.
Our experiments shed light on the interaction between pruning and quantization, where we show that pruning before quantization is beneficial. The elimination of unimportant weight prevents those weights from inducing noise in the quantization process, which preserves accuracy.
The proposed system transforms complex stereo depths estimation networks into more hardware friendly ones with near $99\%$ of memory reduction and $99.97\%$ of hardware cost reduction. The processed networks nearly retain their accuracy both qualitatively (little difference on depth map) and quantitatively (less than $2\%$ of accuracy loss). 

This could soon lead to more efficient hardware, enabling complex models on the edge and on mobile platforms.
{\small
	\bibliographystyle{ieee_fullname}
	\bibliography{egbib}

\begin{thebibliography}{10}\itemsep=-1pt

\bibitem{PSMNet}
Jia-Ren Chang and Yong-Sheng Chen.
\newblock Pyramid stereo matching network.
\newblock In {\em Proceedings of the IEEE Conference on Computer Vision and
  Pattern Recognition}, pages 5410--5418, 2018.

\bibitem{CSPN}
Xinjing Cheng, Peng Wang, and Ruigang Yang.
\newblock Learning depth with convolutional spatial propagation network.
\newblock {\em IEEE transactions on pattern analysis and machine intelligence},
  2019.

\bibitem{chin2020towards}
Ting-Wu Chin, Ruizhou Ding, Cha Zhang, and Diana Marculescu.
\newblock Towards efficient model compression via learned global ranking.
\newblock In {\em Proceedings of the IEEE/CVF Conference on Computer Vision and
  Pattern Recognition}, pages 1518--1528, 2020.

\bibitem{binaryconnect}
Matthieu Courbariaux, Yoshua Bengio, and Jean-Pierre David.
\newblock Binaryconnect: Training deep neural networks with binary weights
  during propagations.
\newblock In {\em Proceedings of the 28th International Conference on Neural
  Information Processing Systems - Volume 2}, NIPS’15, page 3123–3131,
  Cambridge, MA, USA, 2015. MIT Press.

\bibitem{courbariaux2016binarized}
Matthieu Courbariaux, Itay Hubara, Daniel Soudry, Ran El-Yaniv, and Yoshua
  Bengio.
\newblock Binarized neural networks: Training deep neural networks with weights
  and activations constrained to+ 1 or-1.
\newblock {\em arXiv preprint arXiv:1602.02830}, 2016.

\bibitem{guo2020dmcp}
Shaopeng Guo, Yujie Wang, Quanquan Li, and Junjie Yan.
\newblock Dmcp: Differentiable markov channel pruning for neural networks.
\newblock In {\em Proceedings of the IEEE/CVF Conference on Computer Vision and
  Pattern Recognition}, pages 1539--1547, 2020.

\bibitem{han2015deep}
Song Han, Huizi Mao, and William~J. Dally.
\newblock Deep compression: Compressing deep neural networks with pruning,
  trained quantization and huffman coding, 2015.

\bibitem{han2015learn}
Song Han, Jeff Pool, John Tran, and William~J. Dally.
\newblock Learning both weights and connections for efficient neural networks.
\newblock In {\em Proceedings of the 28th International Conference on Neural
  Information Processing Systems - Volume 1}, NIPS’15, page 1135–1143,
  Cambridge, MA, USA, 2015. MIT Press.

\bibitem{resnet}
K. {He}, X. {Zhang}, S. {Ren}, and J. {Sun}.
\newblock Deep residual learning for image recognition.
\newblock In {\em 2016 IEEE Conference on Computer Vision and Pattern
  Recognition (CVPR)}, pages 770--778, June 2016.

\bibitem{resnetv2}
Kaiming He, Xiangyu Zhang, Shaoqing Ren, and Jian Sun.
\newblock Identity mappings in deep residual networks.
\newblock In Bastian Leibe, Jiri Matas, Nicu Sebe, and Max Welling, editors,
  {\em Computer Vision -- ECCV 2016}, pages 630--645, Cham, 2016. Springer
  International Publishing.

\bibitem{he2020LFPC}
Yang He, Yuhang Ding, Ping Liu, Linchao Zhu, Hanwang Zhang, and Yi Yang.
\newblock Learning filter pruning criteria for deep convolutional neural
  networks acceleration.
\newblock In {\em Proceedings of the IEEE/CVF Conference on Computer Vision and
  Pattern Recognition}, pages 2009--2018, 2020.

\bibitem{he2018SoftFP}
Yang He, Guoliang Kang, Xuanyi Dong, Yanwei Fu, and Yi Yang.
\newblock Soft filter pruning for accelerating deep convolutional neural
  networks.
\newblock In {\em International Joint Conference on Artificial Intelligence
  (IJCAI)}, pages 2234--2240, 2018.

\bibitem{he2019GM}
Yang He, Ping Liu, Ziwei Wang, Zhilan Hu, and Yi Yang.
\newblock Filter pruning via geometric median for deep convolutional neural
  networks acceleration.
\newblock In {\em Proceedings of the IEEE Conference on Computer Vision and
  Pattern Recognition (CVPR)}, 2019.

\bibitem{hubara2016quantized}
Itay Hubara, Matthieu Courbariaux, Daniel Soudry, Ran El-Yaniv, and Yoshua
  Bengio.
\newblock Quantized neural networks: Training neural networks with low
  precision weights and activations, 2016.

\bibitem{TFLite}
B. {Jacob}, S. {Kligys}, B. {Chen}, M. {Zhu}, M. {Tang}, A. {Howard}, H.
  {Adam}, and D. {Kalenichenko}.
\newblock Quantization and training of neural networks for efficient
  integer-arithmetic-only inference.
\newblock In {\em 2018 IEEE/CVF Conference on Computer Vision and Pattern
  Recognition}, pages 2704--2713, June 2018.

\bibitem{GANPUISSCC2020}
S. {Kang}, D. {Han}, J. {Lee}, D. {Im}, S. {Kim}, S. {Kim}, and H. {Yoo}.
\newblock 7.4 ganpu: A 135tflops/w multi-dnn training processor for gans with
  speculative dual-sparsity exploitation.
\newblock In {\em 2020 IEEE International Solid- State Circuits Conference -
  (ISSCC)}, pages 140--142, 2020.

\bibitem{cifar10}
Alex Krizhevsky et~al.
\newblock Learning multiple layers of features from tiny images.
\newblock 2009.

\bibitem{li2016ternary}
Fengfu Li, Bo Zhang, and Bin Liu.
\newblock Ternary weight networks, 2016.

\bibitem{li2016pruningfilter}
Hao Li, Asim Kadav, Igor Durdanovic, Hanan Samet, and Hans~Peter Graf.
\newblock Pruning filters for efficient convnets.
\newblock {\em arXiv preprint arXiv:1608.08710}, 2016.

\bibitem{MediaTekDLA}
C. {Lin}, C. {Cheng}, Y. {Tsai}, S. {Hung}, Y. {Kuo}, P.~H. {Wang}, P. {Tsung},
  J. {Hsu}, W. {Lai}, C. {Liu}, S. {Wang}, C. {Kuo}, C. {Chang}, M. {Lee}, T.
  {Lin}, and C. {Chen}.
\newblock 7.1 a 3.4-to-13.3tops/w 3.6tops dual-core deep-learning accelerator
  for versatile ai applications in 7nm 5g smartphone soc.
\newblock In {\em 2020 IEEE International Solid- State Circuits Conference -
  (ISSCC)}, pages 134--136, 2020.

\bibitem{luo2020residual_limited}
Jian-Hao Luo and Jianxin Wu.
\newblock Neural network pruning with residual-connections and limited-data.
\newblock In {\em Proceedings of the IEEE/CVF Conference on Computer Vision and
  Pattern Recognition}, pages 1458--1467, 2020.

\bibitem{luo2017thinet}
Jian-Hao Luo, Jianxin Wu, and Weiyao Lin.
\newblock Thinet: A filter level pruning method for deep neural network
  compression.
\newblock In {\em Proceedings of the IEEE international conference on computer
  vision}, pages 5058--5066, 2017.

\bibitem{KITTI2015C}
Moritz Menze, Christian Heipke, and Andreas Geiger.
\newblock Joint 3d estimation of vehicles and scene flow.
\newblock In {\em ISPRS Workshop on Image Sequence Analysis (ISA)}, 2015.

\bibitem{KITTI2015J}
Moritz Menze, Christian Heipke, and Andreas Geiger.
\newblock Object scene flow.
\newblock {\em ISPRS Journal of Photogrammetry and Remote Sensing (JPRS)},
  2018.

\bibitem{Importance}
Pavlo Molchanov, Arun Mallya, Stephen Tyree, Iuri Frosio, and Jan Kautz.
\newblock Importance estimation for neural network pruning.
\newblock In {\em Proceedings of the IEEE Conference on Computer Vision and
  Pattern Recognition}, pages 11264--11272, 2019.

\bibitem{molchanov2016pruning}
Pavlo Molchanov, Stephen Tyree, Tero Karras, Timo Aila, and Jan Kautz.
\newblock Pruning convolutional neural networks for resource efficient
  inference, 2016.

\bibitem{pytorch}
Adam Paszke, Sam Gross, Soumith Chintala, Gregory Chanan, Edward Yang, Zachary
  DeVito, Zeming Lin, Alban Desmaison, Luca Antiga, and Adam Lerer.
\newblock Automatic differentiation in pytorch.
\newblock 2017.

\bibitem{rastegari2016xnor}
Mohammad Rastegari, Vicente Ordonez, Joseph Redmon, and Ali Farhadi.
\newblock Xnor-net: Imagenet classification using binary convolutional neural
  networks.
\newblock In {\em European conference on computer vision}, pages 525--542.
  Springer, 2016.

\bibitem{imagenet}
Olga Russakovsky, Jia Deng, Hao Su, Jonathan Krause, Sanjeev Satheesh, Sean Ma,
  Zhiheng Huang, Andrej Karpathy, Aditya Khosla, Michael Bernstein, Alexander
  Berg, and Li Fei-Fei.
\newblock Imagenet large scale visual recognition challenge.
\newblock {\em International Journal of Computer Vision}, 115, 09 2014.

\bibitem{nyuv2}
Nathan Silberman, Derek Hoiem, Pushmeet Kohli, and Rob Fergus.
\newblock Indoor segmentation and support inference from rgbd images.
\newblock In Andrew Fitzgibbon, Svetlana Lazebnik, Pietro Perona, Yoichi Sato,
  and Cordelia Schmid, editors, {\em Computer Vision -- ECCV 2012}, pages
  746--760, Berlin, Heidelberg, 2012. Springer Berlin Heidelberg.

\bibitem{vgg}
Karen Simonyan and Andrew Zisserman.
\newblock Very deep convolutional networks for large-scale image recognition,
  2014.

\bibitem{ebp}
Daniel Soudry, Itay Hubara, and Ron Meir.
\newblock Expectation backpropagation: Parameter-free training of multilayer
  neural networks with continuous or discrete weights.
\newblock In {\em Proceedings of the 27th International Conference on Neural
  Information Processing Systems - Volume 1}, NIPS’14, page 963–971,
  Cambridge, MA, USA, 2014. MIT Press.

\bibitem{INQ}
Aojun Zhou, Anbang Yao, Yiwen Guo, Lin Xu, and Yurong Chen.
\newblock Incremental network quantization: Towards lossless cnns with
  low-precision weights.
\newblock {\em arXiv preprint arXiv:1702.03044}, 2017.

\bibitem{zhung2018}
Bohan Zhuang, Chunhua Shen, Mingkui Tan, Lingqiao Liu, and Ian Reid.
\newblock Towards effective low-bitwidth convolutional neural networks.
\newblock pages 7920--7928, 06 2018.

\end{thebibliography}
}

\clearpage
\appendix
\section{Training Settings for ResNet}
For CIFAR-10, we followed the implementation of \cite{resnet}. In the ILSVRC-2012 experiments, we used the default parameter settings of \cite{resnet}. The pre-trained models are from PyTorch~\cite{pytorch}'s TorchVision library. Data augmentation strategies are the same as in PyTorch~\cite{pytorch}'s official examples. For pruning the pre-trained model, we pruned for $100$ epochs, used a learning rate of $0.001$ and reduce the learning rate by half after $50$ epochs. For pruning the model from scratch on CIFAR-10, we use the normal training schedule without an additional fine-tune process. 

\section{Pruning Results with PSM-Net}
\begin{figure}[h!]
\centering
\subfloat[Accuracy]{
\begin{minipage}[c]{0.46\textwidth} 
\includegraphics[width=1\textwidth]{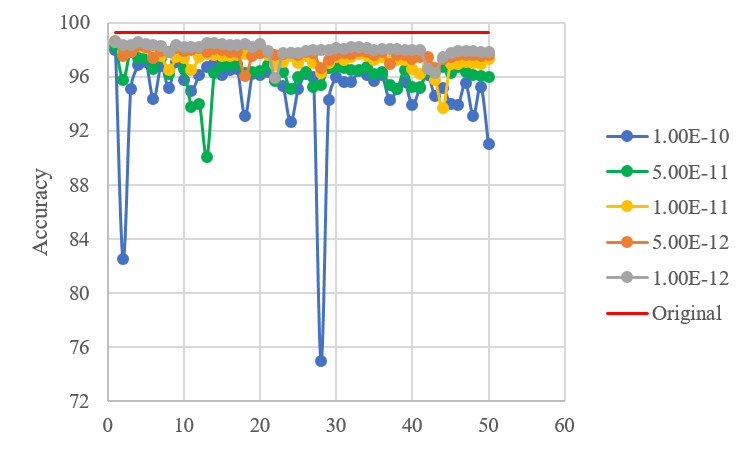}
\end{minipage}
}

\subfloat[Sparsity]{
\begin{minipage}[c]{0.46\textwidth} 
\includegraphics[width=1\textwidth]{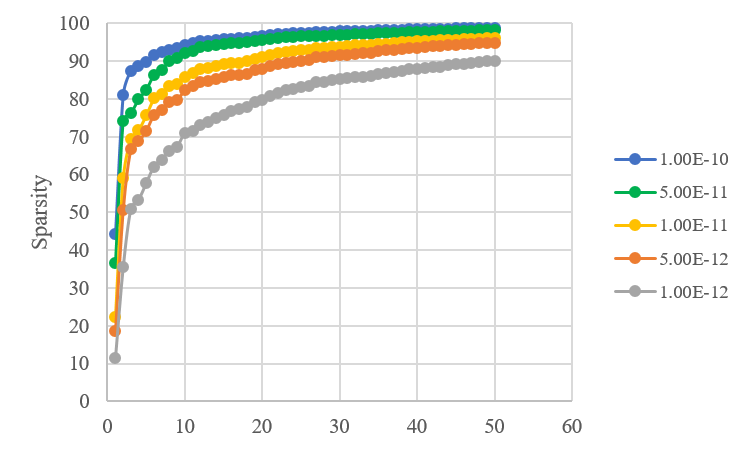}
\end{minipage}
}
\caption{Pruning Results of Different Threshold of (a) the accuracy and (b) the achieved weight sparsity }
\label{fig:Pruning}
\end{figure}

The sparsity can be controlled by changing the pruning threshold. 
If the threshold is set higher, more weights are pruned and more gates set to $0$. 
We tested five different thresholds using our proposed pruning. 
The results in Fig.~\ref{fig:Pruning} clearly indicate that our Taylor Score importance pruning technique can continuously increase the sparsity while the fine-tuning in each epoch can effectively compensate the accuracy loss. 
As a result, we can obtain a pruned model with $97\%$ sparsity and only $2\%$ of accuracy loss, effectively compressing the large stereo estimation neural network. 
We choose $1e-11$ as pruning threshold and use this threshold value in our other experiments.

\begin{table*}[h!]
	\centering
	\small
	\caption{Different Quantization Step}
	\label{table:accimp}
	\begin{tabular}{|c|c|c|c|c|c|c|c|c|}
		\hline
		Steps    & 0        & 0.5      & 0.625    & 0.75     & 0.875    & 0.9      & 0.95     & 1        \\ \hline
		Accuracy & 97.39 & 96.72 & 96.56 & 96.15 & 95.59 & 95.79 & 95.21 & 94.42 \\ \hline
		Steps    & 0        & 0.5      & 0.75     & 0.875    & 0.9      & 0.95     & 0.975    & 1        \\ \hline
		Accuracy & 97.41  & 96.72 & 96.21 & 95.28  & 94.93 & 94.83 & 94.28 & 94.05 \\ \hline
	\end{tabular}
\end{table*}
\begin{table*}[h!]
	\centering
	\footnotesize
	\caption{Increase Quantization Steps}
	\label{table:moreQstep}
	\begin{tabular}{ccccccccc}
		\hline
		\multicolumn{1}{|c|}{Step}    & \multicolumn{1}{c|}{0}        & \multicolumn{1}{c|}{0.5}      & \multicolumn{1}{c|}{0.625}    & \multicolumn{1}{c|}{0.75}     & \multicolumn{1}{c|}{0.875}    & \multicolumn{1}{c|}{0.9}      & \multicolumn{1}{c|}{0.95}     & \multicolumn{1}{c|}{1}        \\ \hline
		\multicolumn{1}{|c|}{Accuracy} & \multicolumn{1}{c|}{97.39} & \multicolumn{1}{c|}{96.72} & \multicolumn{1}{c|}{96.56} & \multicolumn{1}{c|}{96.15} & \multicolumn{1}{c|}{95.6} & \multicolumn{1}{c|}{95.79} & \multicolumn{1}{c|}{95.21} & \multicolumn{1}{c|}{\textbf{94.42}} \\ \hline
		&                               &                               &                               &                               &                               &                               &                               &                               \\ \cline{1-8}
		\multicolumn{1}{|c|}{Step}     & \multicolumn{1}{c|}{0}        & \multicolumn{1}{c|}{0.5}      & \multicolumn{1}{c|}{0.625}    & \multicolumn{1}{c|}{0.75}     & \multicolumn{1}{c|}{0.8}      & \multicolumn{1}{c|}{0.825}    & \multicolumn{1}{c|}{0.875}    &                               \\ \cline{1-8}
		\multicolumn{1}{|c|}{Sparsity} & \multicolumn{1}{c|}{95.6}     & \multicolumn{1}{c|}{74.08}    & \multicolumn{1}{c|}{80.11}    & \multicolumn{1}{c|}{86.12}    & \multicolumn{1}{c|}{88.54}    & \multicolumn{1}{c|}{89.75}    & \multicolumn{1}{c|}{92.16}    &                               \\ \cline{1-8}
		\multicolumn{1}{|c|}{Accuracy} & \multicolumn{1}{c|}{97.37}    & \multicolumn{1}{c|}{96.69}    & \multicolumn{1}{c|}{96.53}    & \multicolumn{1}{c|}{96.17}    & \multicolumn{1}{c|}{96.10}    & \multicolumn{1}{c|}{96.24}    & \multicolumn{1}{c|}{95.88}    &                               \\ \cline{1-8}
		\multicolumn{1}{|c|}{Step}     & \multicolumn{1}{c|}{0.925}    & \multicolumn{1}{c|}{0.95}     & \multicolumn{1}{c|}{0.975}    & \multicolumn{1}{c|}{0.9875}   & \multicolumn{1}{c|}{0.99}     & \multicolumn{1}{c|}{0.995}    & \multicolumn{1}{c|}{1}        &                               \\ \cline{1-8}
		\multicolumn{1}{|c|}{Sparsity} & \multicolumn{1}{c|}{94.56}    & \multicolumn{1}{c|}{95.76}    & \multicolumn{1}{c|}{96.97}    & \multicolumn{1}{c|}{97.57}    & \multicolumn{1}{c|}{97.69}    & \multicolumn{1}{c|}{97.93}    & \multicolumn{1}{c|}{98.17}    &                               \\ \cline{1-8}
		\multicolumn{1}{|c|}{Accuracy} & \multicolumn{1}{c|}{95.92}    & \multicolumn{1}{c|}{95.68}    & \multicolumn{1}{c|}{95.87}    & \multicolumn{1}{c|}{95.61}    & \multicolumn{1}{c|}{95.25}    & \multicolumn{1}{c|}{95.4}     & \multicolumn{1}{c|}{\textbf{95.05}}    &                               \\ \cline{1-8}
	\end{tabular}
\end{table*}

\section{Different Weight Bits}
\begin{figure}[h!]
	\centering
	\begin{tabular}{l}
		\includegraphics[scale=0.395]{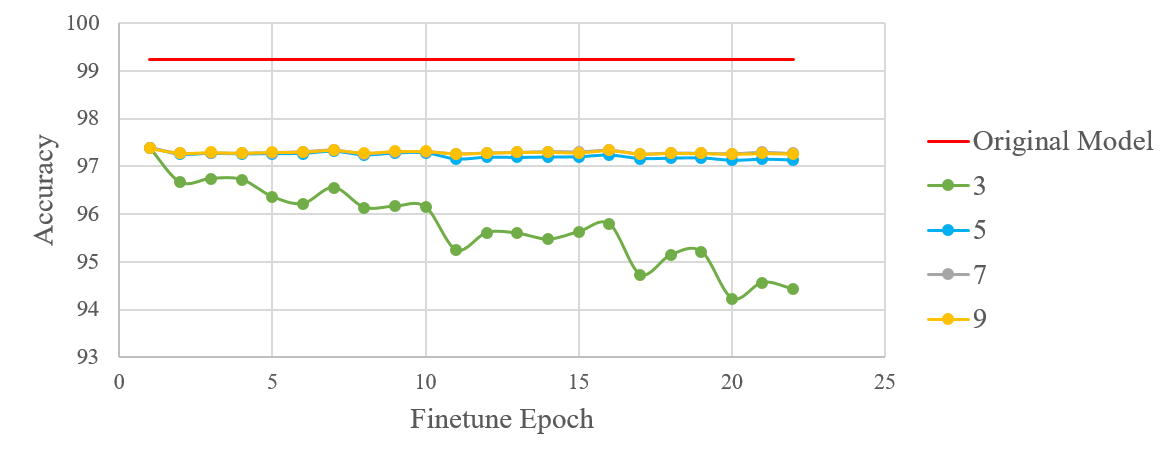} \\
		\qquad \qquad \qquad \qquad \ \ (a) PSM-Net \\
		\includegraphics[scale=0.395]{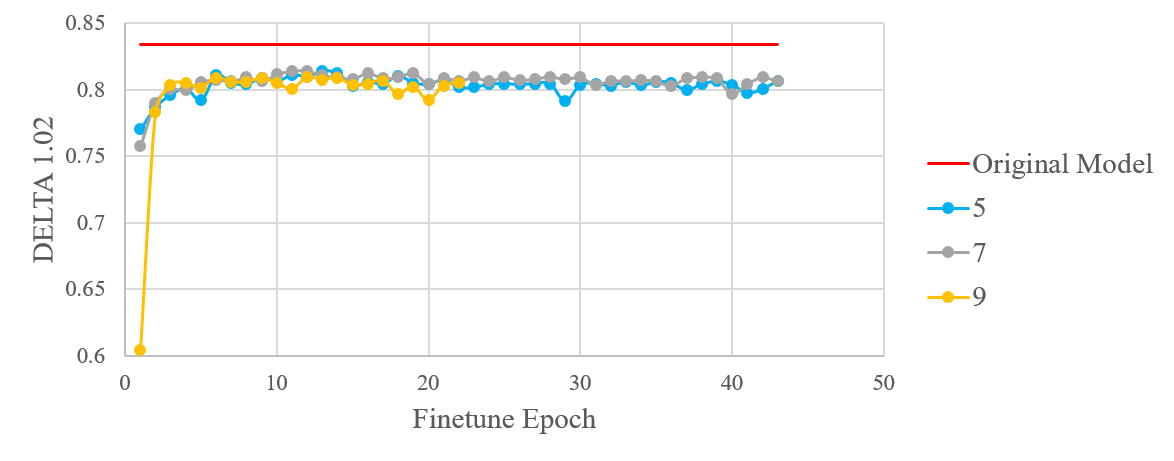}\\
		\qquad \qquad \qquad \qquad \ \ (b) CSPN\\
	\end{tabular}
	\caption{Accuracy comparison with different weight bits with various stereo neural networks (a) PSM-Net~\cite{PSMNet} and (b) CSPN~\cite{CSPN}. }
	\label{fig:diff_WB}
\end{figure}

In this section, we discuss different quantization weight bits for different kinds of input stereo neural networks.
Fig.~\ref{fig:diff_WB}(a) shows the accuracy drop of PSM-Net for different weight bits after the quantization process. 
It is obvious that the performance drop increases as the number of weight bits decreases, but we can observe that for weight bits $9$, $7$ and $5$ the validation results of the network vary only little. 
The results indicate that by using our technique, we can map the weight to a limited number of powers of $2$ with little performance drop. 
In the case of $5$ weight bit, there are only eight choices each for positive and negative numbers.
Even with extremely low weight bits, 3 bits, the performance drop is still tolerable and visually unrecognizable by humans. 
Table~\ref{table:diff_WB} suggested that choosing smaller weight bits helps preserve the sparsity obtained from the pruning phase. 
A sparsity-accuracy trade-off can also be observed from weight bit $5$ and $3$, which turn out to be better choices comparing to higher weight bits such as $9$ and $7$.  

Similar to PSM-Net, we explore different quantization weight bits for CSPN. 
To evaluate the best possible results, we use Taylor Score as the weight partition metric because it performs best among three methods (absolute value, Taylor and random) on CSPN. 
Since higher weight bits require less epochs to fine-tune, we reduced the fine-tune epochs when the weight bit is 9. 
As the result shown in Fig. ~\ref{fig:diff_WB}(b), setting weight bits to $5$ does not decrease the performance much. 
In comparison with higher weight bits such as $9$ and $7$, the model performance is similar. 
Thus, by applying our technique on a larger stereo matching network like CSPN, we can also obtain a highly sparse and fully-quantized model with negligible performance loss. 
The final sparsity is shown in Table ~\ref{table:diff_WB}. 
On PSM-Net we can further reduce the weight bits to $3$, but since CSPN is a much more complicated network with significantly more parameters than PSM-Net, extremely low weight bits ($3$) would severely harm the network performance. 
From the perspective of a VLSI (hardware) implementation, the complexity of weight bits $5$ and $3$ in our proposed system are similar. 
Since our technique quantizes all the weights to powers of two, it requires only simple shifters instead of complicated MACs.

\begin{table}[h!]
	\centering
	\small
	\caption{Sparsity ($\%$) of Different Weight Bits on Different Input Neural Networks}
	\label{table:diff_WB}
	\begin{tabular}{ccccc}
		\toprule
		& 9       & 7       & 5   & 3     \\ \midrule
		PSM-Net & 72.813 & 72.879 & 95.978 & 98.17              
		\\ \midrule
		CSPN & 92.9333 & 92.9039 & 93.7271 & NA \\ \bottomrule
	\end{tabular}
\end{table}

\section{Accuracy Improvement on PSM-Net}
In spite of a negligible performance drop, we still conduct experiments to search for potential ways to retain model performance under extremely low weight bit (weight bit $=3$). Table~\ref{table:accimp} shows that changing the distribution of the quantization step alone cannot improve performance. The early step is also crucial to the performance. Table~\ref{table:moreQstep} shows that to improve performance, the quantization steps must increase on both, early steps and late steps. The upper table is the original quantization step and the bottom one is the increased quantization step. The results further indicate that the fine-tune process of our pipeline is effective. If more steps are adopted, which means that more retraining are occurring, the performance drop can be reduced further. 

\section{Supplementary Graphic}
\begin{figure}[h!]
	\centering
	\includegraphics[scale=0.3]{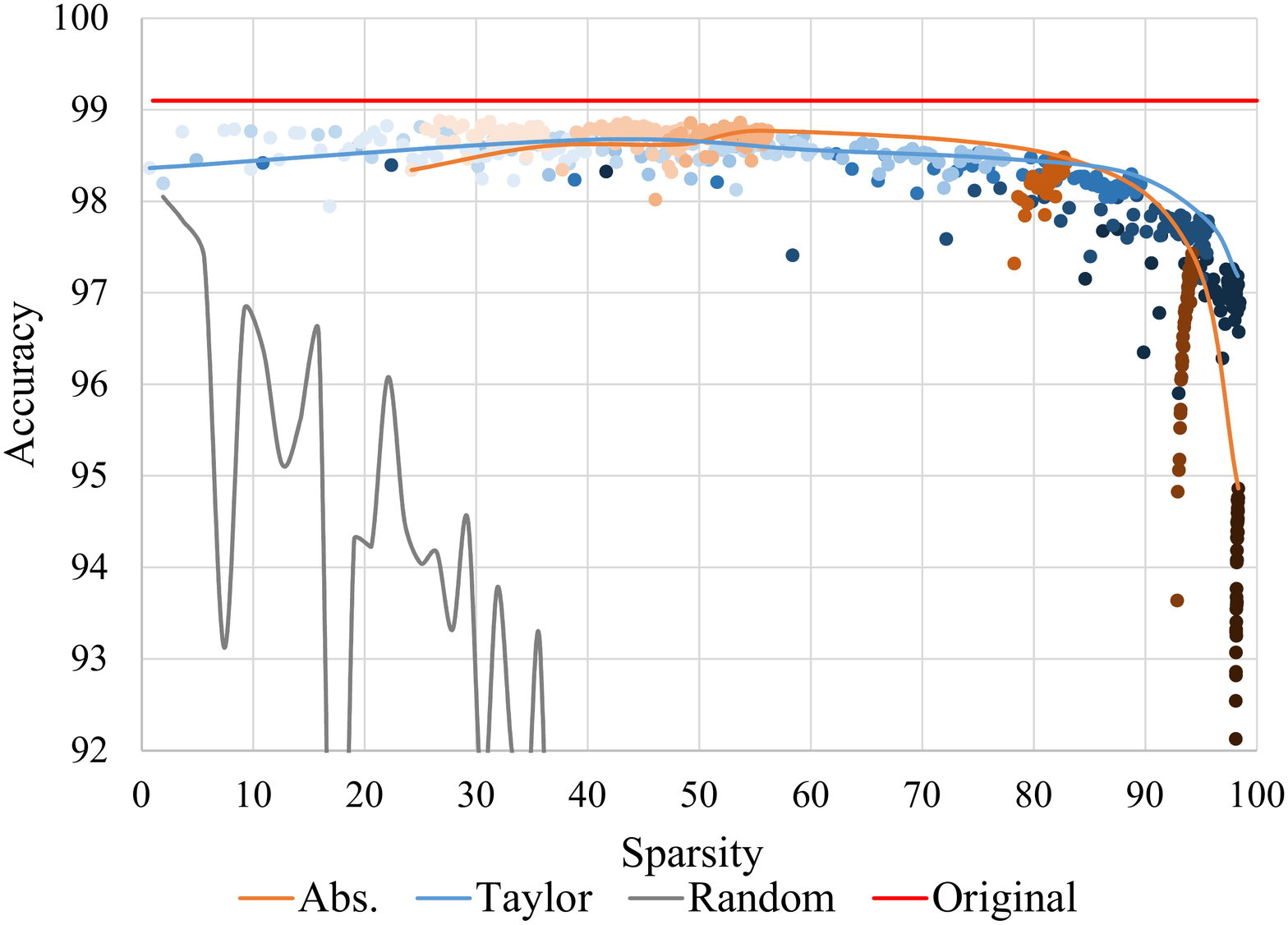}
	\caption{Full axis of accuracy and sparsity of Fig.~\ref{fig:criteria_2}. Dots in blue shades are Taylor Score results after each epoch, while dots in orange shades are results for the absolute value criterion. Darker dots mean higher threshold. 
	}
	\label{fig:criteria}
\end{figure}

Fig.~\ref{fig:criteria} is the full graph of accuracy and sparsity of Fig.~\ref{fig:criteria_2}.

\clearpage
\section{More Visualization Results}
\begin{figure}[htbp]
	\centering
	\includegraphics[scale=0.5]{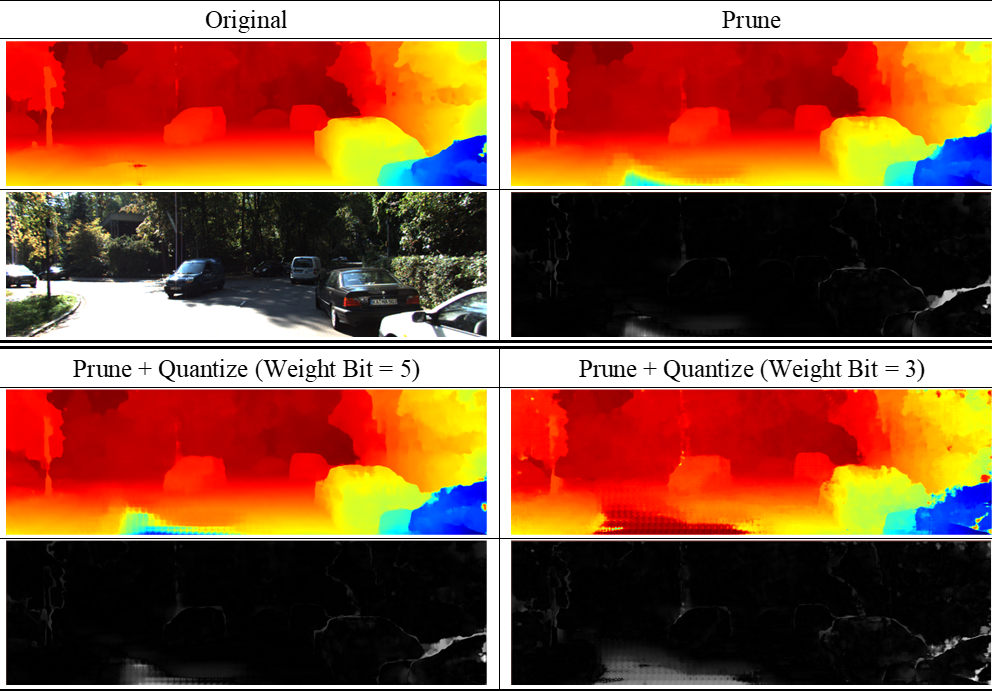}
	\caption{}
	\label{fig:qual_appendix1}
\end{figure}

\begin{figure}[htbp]
	\centering
	\includegraphics[scale=0.5]{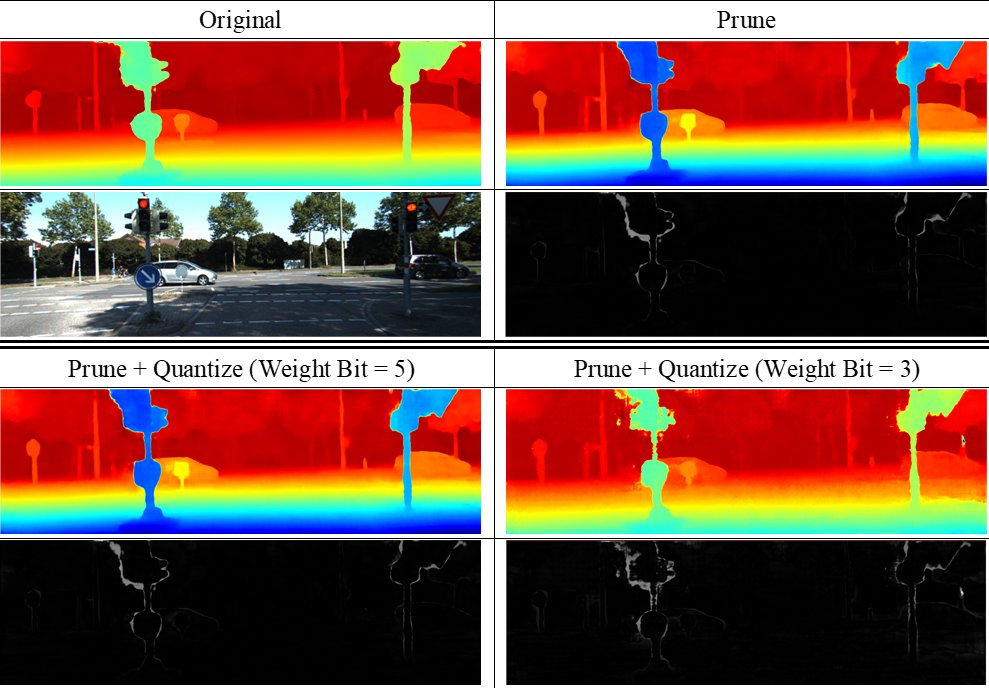}
	\caption{}
	\label{fig:qual_appendix2}
\end{figure}

\begin{figure}[htbp]
	\centering
	\includegraphics[scale=0.4]{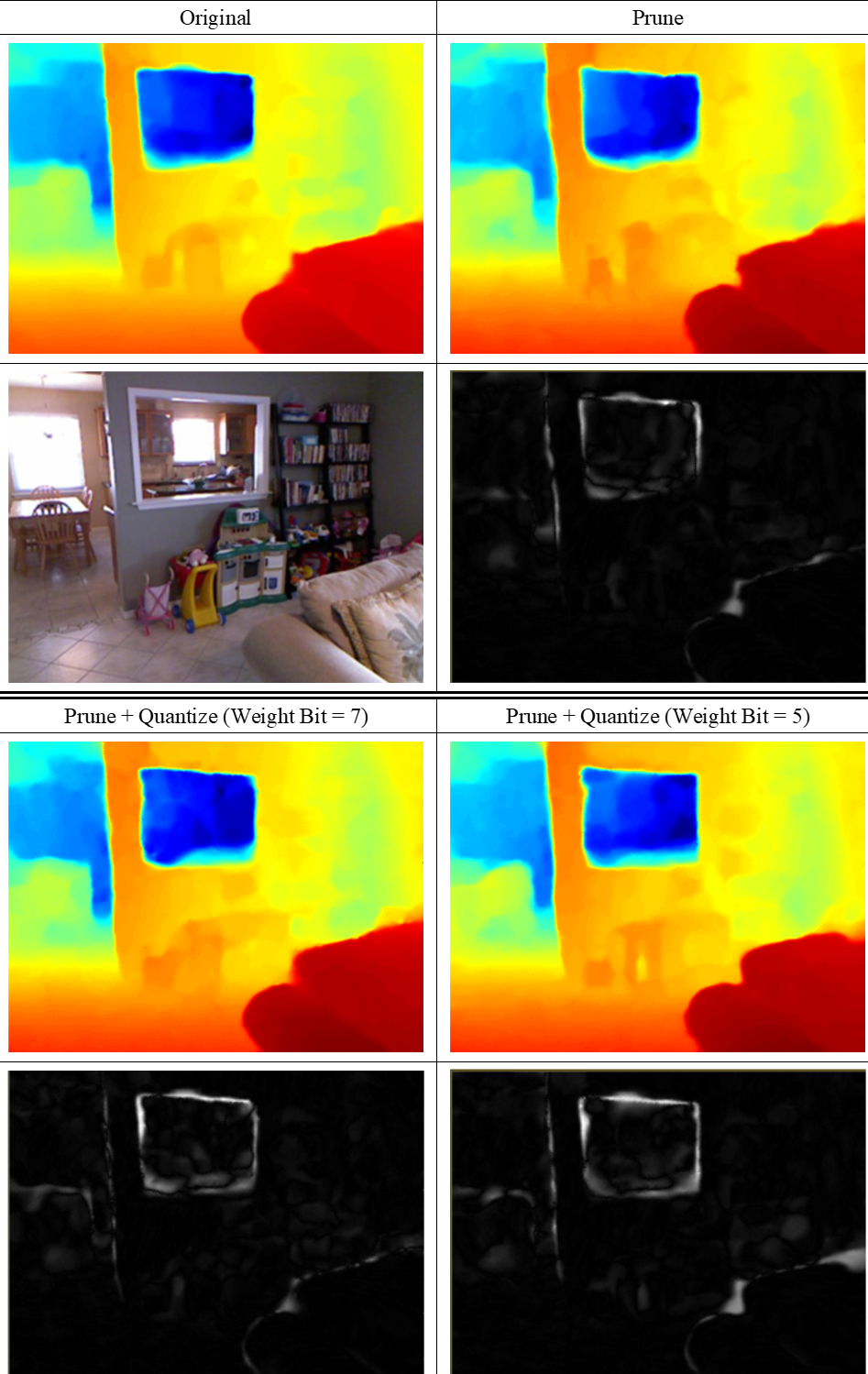}
	\caption{}
	\label{fig:qual_appendix_cspn2}
\end{figure}

\begin{figure}[htbp]
	\centering
	\includegraphics[scale=0.4]{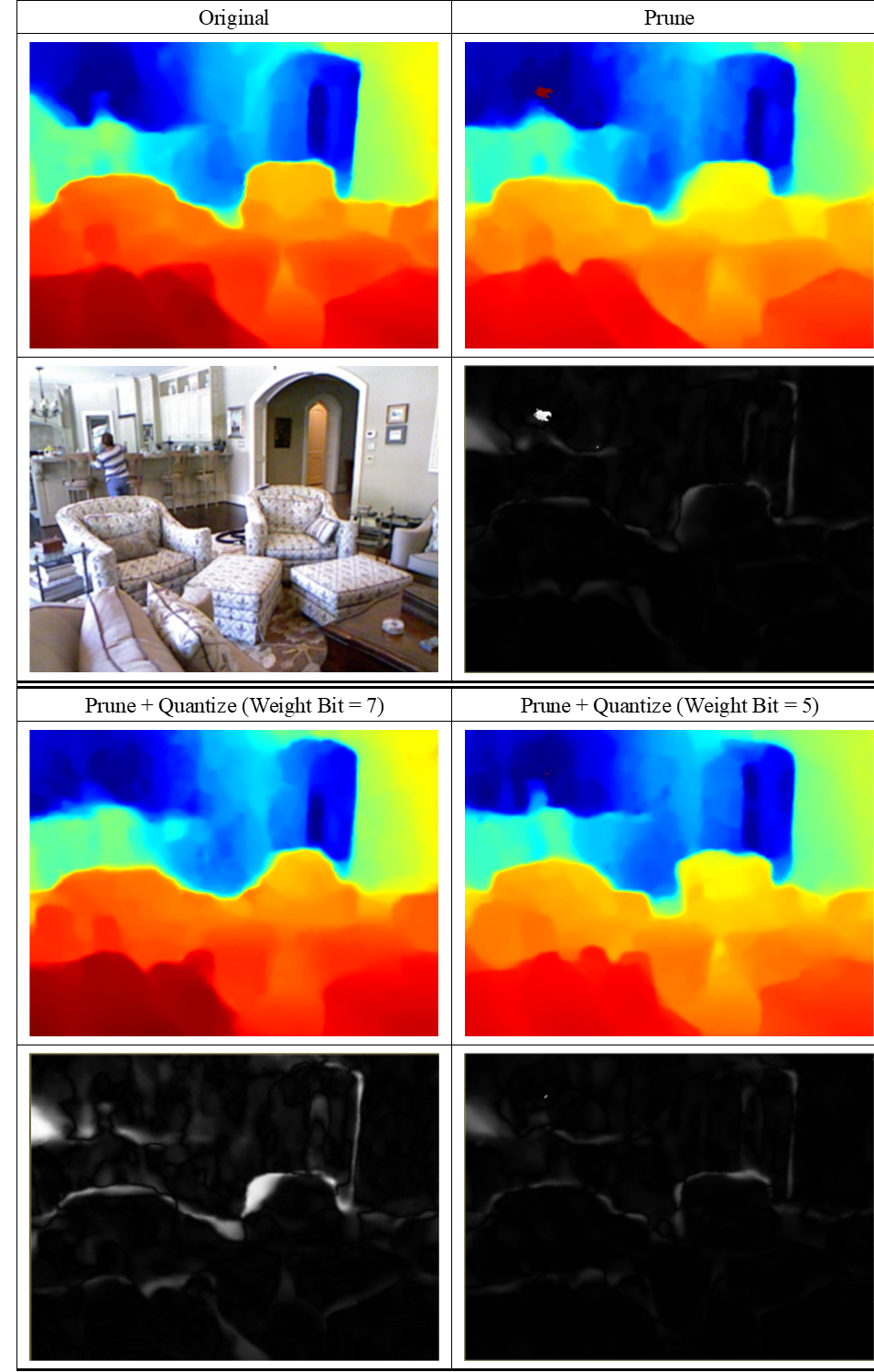}
	\caption{}
	\label{fig:qual_appendix_cspn3}
\end{figure}

We provide more depth map visualization results for both PSMNet and CSPN.
For PSMNet, the upper image of each of the pairs shows the disparity map obtained in different settings (original model, pruned model, pruned and quantized to 5 weigh bits, pruned and quantized to 3 weight bits). 
For CSPN the upper image of each of the pairs shows the disparity map obtained in different settings (original model, pruned model, pruned and quantized to 7 weigh bits, pruned and quantized to 3 weight bits).
The bottom image shows the disparity difference between the original model and the predicted results. Brightness indicates difference. Fig.~\ref{fig:qual_appendix1}~\ref{fig:qual_appendix2} are extracted from KITTI2015 dataset and processed by PSMNet; Fig.~\ref{fig:qual_appendix_cspn2}~\ref{fig:qual_appendix_cspn3} are extracted from NYU depth V2 dataset and processed by CSPN.

\end{document}